\documentclass[10pt,letterpaper]{uconn-minds}

\usepackage{amsmath,amssymb}
\usepackage{float,xspace,multirow,enumitem,array,adjustbox,wrapfig,needspace}
\graphicspath{{figures/}}
\definecolor{promptblue}{HTML}{0072B2}
\definecolor{promptfill}{HTML}{E6F2FA}
\definecolor{groupgray}{HTML}{EFEFEF}
\renewcommand{\arraystretch}{1.08}

\newcommand{\finishwrap}{%
  \par
  \ifnum\value{WF@wrappedlines}>1
    \vspace{\dimexpr\value{WF@wrappedlines}\baselineskip-\baselineskip\relax}%
  \fi
  \WFclear}
\newcommand{\uncertaintyfont}{\fontsize{6.5}{7.5}\selectfont}
\newcommand{\tabfont}{\fontsize{8.4}{9.8}\selectfont\let\scriptsize\uncertaintyfont}
\newcommand{\pmse}[1]{{\uncertaintyfont$\pm$#1}}
\newcommand{\priv}{$^{\dagger}$}
\newcommand{\bench}{UMAMI\xspace}
\newcommand{\benchlong}{\textbf{U}nified \textbf{M}ultimodal \textbf{A}ssessment of \textbf{M}eals and \textbf{I}ntake\xspace}
\newcommand{\tvqa}{Visual Question Answering\xspace}
\newcommand{\trec}{Personalized Food Recommendation\xspace}
\newcommand{\tsub}{Condition-Aware Food Substitution\xspace}

\title{UMAMI: A Multimodal Benchmark Dataset for\\Evidence-Grounded Dietary Decisions}
\author{Safal Thapaliya$^{1}$, Jiatan Huang$^{1}$, Debapriya Chatterjee$^{1}$\\
Yanfang Ye$^{2}$, Chuxu Zhang$^{1,\dagger}$}
\affiliation{$^{1}$University of Connecticut\\
$^{2}$University of Notre Dame\\
$^{\dagger}$Corresponding author}
\runningtitle{UMAMI}
\date{}
\hypersetup{
  pdftitle={UMAMI: A Multimodal Benchmark Dataset for Evidence-Grounded Dietary Decisions},
  pdfauthor={Safal Thapaliya, Jiatan Huang, Debapriya Chatterjee, Yanfang Ye, Chuxu Zhang},
  pdfsubject={Multimodal dietary decision benchmark},
  pdfkeywords={UMAMI, multimodal benchmark, food grounding, dietary decisions}
}
\paperabstract{%
Dietary decisions require identifying a food's nutrient content and considering individual dietary needs, but recognizing food from an image alone does not provide this information. Visually similar foods can have different nutrient profiles, so selecting the wrong food record can lead to incorrect decisions even when nutrient rules are applied correctly. Furthermore, satisfying nutrient constraints does not necessarily establish a food's relevance to a person's intake. Evaluating these challenges together requires connecting food images, nutrient records, and individual intake histories. To support this evaluation, we introduce UMAMI (\textbf{U}nified \textbf{M}ultimodal \textbf{A}ssessment of \textbf{M}eals and \textbf{I}ntake), a multimodal dataset and benchmark linking independently collected food images with nutrient profiles and respondent health and intake records. Using this dataset, we construct three diagnostic tasks covering visual nutrient question answering, personalized food recommendation, and condition-aware food substitution. Paired instances and input controls examine how visual identity and respondent context affect decisions. Our experiments identify errors in both food-record selection and nutrient-rule application, while showing that higher rule compliance does not necessarily improve alignment with subsequent recorded intake. These findings highlight the need to evaluate how multimodal systems select and use structured evidence for dietary decisions. The dataset, code, and dashboard are available at \url{https://umami-benchmark.github.io}.
}

\begin{document}
\raggedbottom
\maketitle
\thispagestyle{firstpage}
\section{Introduction}
\label{sec:intro}

\Needspace{195pt}
\begin{wrapfigure}{r}{0.48\textwidth}
\centering
\includegraphics[width=\linewidth]{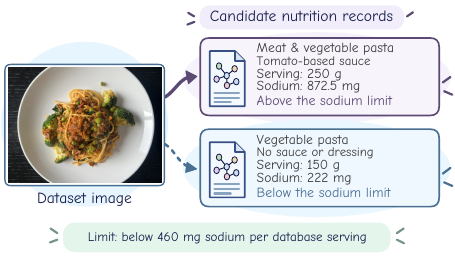}
\captionsetup{justification=raggedright,singlelinecheck=false}
\caption{Selecting the wrong food record reverses the sodium-limit decision.
Solid arrows: correct grounding, and dashed arrows: incorrect grounding.
}
\label{fig:food-record-example}
\end{wrapfigure}

Dietary choices affect health, but whether a food fits an individual's needs depends on both its nutrient content and the person's dietary requirements~\citep{gbd2019diet,berry2020postprandial}.
Images provide a practical way to present food choices~\citep{cordeiro2015rethinking,boushey2017mobile}.
Accessing these choices first requires linking the depicted foods to nutrient records so that dietary constraints can be checked.
Among foods that satisfy these constraints, however, nutrient content alone does not indicate which options are relevant to a particular person's intake.
This decision requires context from the person's dietary history.

Linking an image to the appropriate nutrient record is
challenging because visually similar foods can differ in nutrient
content~\citep{yan2025dietai24,lin2023integration}.
In Figure~\ref{fig:food-record-example}, pasta with sauce exceeds the sodium
limit, whereas vegetable pasta without sauce remains below it, at their
database-defined servings. Recognizing ``pasta'' is therefore insufficient:
selecting the wrong record can reverse the decision even when the nutrient
rule is applied correctly.
Evaluating this process requires separating errors in record selection from errors in using the selected values.

\begin{figure}[!t]
\centering
\includegraphics[width=\linewidth]{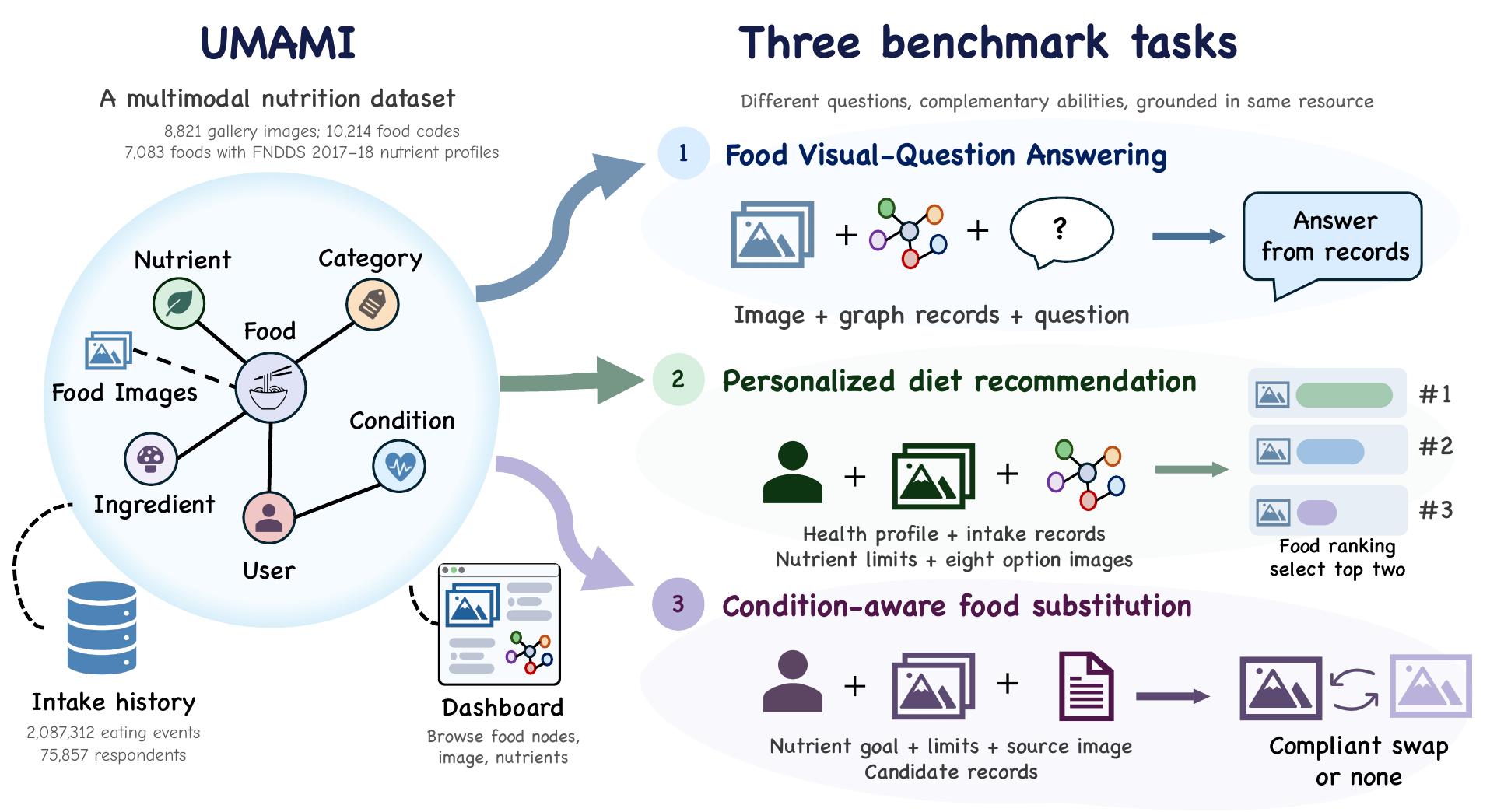}
\caption{\bench{} connects food images with nutrient records, intake histories, and health
conditions (left). The three tasks (right) evaluate nutrient question answering, personalized
ranking, and constrained substitution, including rejection when no candidate qualifies.
Images depict food types rather than respondents' meals.}
\label{fig:overview}
\end{figure}

Existing evaluations address different parts of this process.
Graph-based nutrition benchmarks link nutrient and respondent records but provide food identities directly~\citep{zhang2024nutrition,glenbench2026}.
Visual food benchmarks typically lack respondent-level context~\citep{thames2021nutrition5k,li2024foodieqa,jin2026diningbench}, and image-to-nutrient grounding studies omit individual intake histories~\citep{yan2025dietai24,hu2025healthmfqa}.
These complementary gaps motivate an evaluation that follows the same evidence from visual food identification through nutrient-rule decisions and intake-based selection.

To support this evaluation, we introduce \bench{} (\benchlong{}), which connects an initial corpus of 20,492 food images to food-code records and health and intake data from 75,857 National Health and Nutrition Examination Survey (NHANES) respondents ~\citep{ahluwalia2016nhanes}.
The nutrient profiles are available for a subset of food codes, and task instances require the nutrient fields used by their rules.
The images were collected independently of the survey and depict food types rather than meals consumed by respondents. Food codes connect these images to nutritional records and reported intake without implying that the images document individual consumption.
Three diagnostic tasks assess complementary decisions (Figure~\ref{fig:overview}): \tvqa{} answers nutrient questions about depicted foods, \trec{} selects relevant foods under dietary constraints, and \tsub{} identifies replacements that meet a nutrient goal.

\bench{} provides a diagnostic evaluation of image-grounded dietary
decisions. Its linked records allow us to distinguish three capabilities:
selecting the appropriate food record, correctly applying nutrient rules,
and ranking compliant foods according to intake-derived relevance.
Paired instances examine cases in which changes in food identity or
respondent context change the correct answer. Input controls further
examine whether errors arise from record selection or use of the
selected evidence.
Our contributions are as follows:

\begin{itemize}[leftmargin=1.2em,itemsep=1pt,topsep=2pt]
\item We introduce \bench{}, a multimodal dataset linking food images with
nutrient profiles and respondent health and intake records.
\item We provide a benchmark framework with three paired tasks and input
controls to assess food identification, nutrient-rule application, and
intake-based recommendation.
\item  Our experiments show that some models fail to apply nutrient rules even with correct food records, and that higher rule compliance does not always coincide with higher accuracy against intake-derived relevance labels.
\end{itemize}

\section{Dataset Construction}
\label{sec:overview}
\label{sec:background}
\label{sec:resource}

\Needspace{250pt}
\begin{wraptable}{r}{0.54\textwidth}
\centering\tabfont
\begin{adjustbox}{max width=\linewidth}
\begin{tabular}{@{}lr@{}}
\toprule
\textbf{Resource} & \textbf{Count} \\
\midrule
\rowcolor{groupgray}\multicolumn{2}{@{}l}{\textit{Survey and graph}} \\
Respondents & 75{,}857 \\
Reliable eating events & 2{,}087{,}312 \\
Food codes / with FNDDS 2017--18 profiles & 10{,}214 / 7{,}083 \\
\rowcolor{groupgray}\multicolumn{2}{@{}l}{\textit{Initial image corpus}} \\
Images / image--food links & 20{,}492 / 30{,}274 \\
Gallery images & 8{,}821 \\
Initial queries / cross-source & 10{,}179 / 9{,}202 \\
Images for file distribution & 2{,}040 \\
\rowcolor{groupgray}\multicolumn{2}{@{}l}{\textit{Task instances: test}} \\
VQA & 2{,}708 \\
Recommendation & 406 \\
Substitution & 1{,}220 \\
\bottomrule
\end{tabular}
\end{adjustbox}
\captionsetup{justification=raggedright,singlelinecheck=false}
\caption{We summarize the initial resource and test instances. The separate test-query extension contains
8{,}428 images. Substitution uses a separate query pool.}
\label{tab:dataset-stats}
\end{wraptable}

Food codes provide the connection between visual food identity, nutrient
content, and personal intake. In NHANES dietary recalls, these codes link
reported foods to descriptions, nutrient profiles, and serving weights in
the Food and Nutrient Database for Dietary Studies (FNDDS). Associating
independently collected images with the same codes makes it possible to
study how identifying a depicted food affects subsequent decisions.
We refer to selecting its associated record as \emph{food grounding}.
To connect these decisions with personal context, we combine recalls with
examinations, laboratory measurements, and
questionnaires~\citep{ahluwalia2016nhanes,moshfegh2008ampm}.
Reconciling releases and retaining reliable recalls yields
2{,}087{,}312 eating events, with 64{,}859 respondents providing two recalls.
The graph links foods to images, ingredients, What We Eat in America
(WWEIA) categories, and intake records. Survey measurements establish
hypertension, diabetes, high cholesterol, and obesity status, while missing
values remain unknown. A single FNDDS reference keeps nutrient-rule inputs
consistent across survey cycles and tasks.

To obtain visual inputs for these records, we retrieve images using food
descriptions from Recipe1M, Openverse, Wikimedia Commons, Open Food Facts,
and web search. Image--text agreement and safety screening precede a model
panel's assessment of food identity and single-food presentation.
Because these models influence collection, we exclude the panel checkpoints
and CLIP filter from evaluation. Minimum-cost one-to-one assignment then
selects distinct verified gallery images for 8{,}821 foods. Separate query
images serve as task inputs; each must map to one food code and pass size
and duplicate checks. Perceptual hashes and DINOv2 similarity exclude
near-duplicates of gallery images, leaving 10{,}179 initial queries,
including 9{,}202 from sources different from their gallery counterparts.
All tasks require the nutrients used by their rules.
VQA and substitution supply candidate records, while
recommendation retrieves them from the catalog before
checking rules and ranking.

Evaluation separates respondents by survey cycle, duplicate-image groups
by task partition, and food-code prefixes by VQA split. Static catalog
attributes are shared across splits, but behavioral edges and popularity
estimates use only training respondents, excluding evaluation histories.
A test-only extension adds 8{,}428 queries accepted by a stricter later
panel, mostly from web search, and supplies 1{,}954 of the 2{,}392 test
images. Substitution uses a separate query pool
(Appendix~\ref{sec:app-resource}). To assess collection quality, a blinded
human audit examines 500 items. The three complete reviews yield
population-weighted false-acceptance rates of 7.9\%, 10.2\% and 11.7\%, with
mean pairwise Cohen's $\kappa=0.40$ (Appendix~\ref{sec:app-audit}).
These estimates concern the initial collection, excluding the test-only
extension and task labels; agreement with an assigned record does not
independently establish its visual validity (Appendix~\ref{sec:app-audit}).

\section{Benchmark Design}
\label{sec:design}
\label{sec:design:steps}
\label{sec:design:tasks}
\label{sec:design:boundary}
\label{sec:design:gold}

\bench{} is a diagnostic benchmark for image-grounded dietary decisions, with three tasks evaluating complementary capabilities. \tvqa{} evaluates whether models can match food images to the correct records and use their nutrient values to answer questions. \trec{} evaluates whether models can satisfy nutrient constraints and rank foods by their relevance to subsequently observed intake. \tsub{} evaluates whether models can identify the source food and apply all nutrient constraints when selecting a replacement. Together, these tasks assess visual grounding, nutrient-rule application, and intake-aligned ranking.

\label{sec:design:pairs}
\label{sec:design:controls}
For VQA and substitution, paired examples share the question and candidate
records but depict different foods that require different answers.
Recommendation pairs provide the same food options and nutrient constraints to respondents with different intake histories and corresponding relevance labels.
These comparisons examine whether predictions appropriately distinguish changes in visual or personal information~\citep{thrush2022winoground,li2024naturalbench}.
Because recommendation option positions are shuffled independently, those pairs do not fully isolate intake-history effects from presentation-order effects.
We further examine evidence use by removing inputs, reordering images, and supplying true food identities ~\citep{chen2023infoseek,mensink2023encyclopedic}
(Appendix~\ref{sec:app-taskdetail}).

\label{sec:design:leaks}
\label{sec:design:server}
Models may also exploit cues unrelated to the intended decision.
We therefore check for potential shortcuts involving public fields, option ordering,
food popularity, and text within images
(Appendix~\ref{sec:app-checks}).
These checks address specific alternative explanations without ruling out shortcuts.

\section{Systems and Evaluation Protocol}
\label{sec:setup}
\label{sec:design:metrics}

\paragraph{Baselines.}
We compare direct vision--language models, image-to-record matching pipelines, graph-based readers, and trained recommendation models to examine different stages of the decision process. These systems differ in their access to structured evidence and their inference procedures.
We therefore use the comparisons to characterize the evaluated configurations rather than to isolate architectural differences across all systems.
Direct models include Qwen3-VL, InternVL, GLM,
and MiMo~\citep{qwen3vl2025,wang2025internvl35,glmv2025glm41v,xiaomi2025mimovl}.
Matching pipelines identify records using DINOv2, SigLIP2, or Qwen3-VL-8B
and then apply the nutrient rules~\citep{oquab2024dinov2,tschannen2025siglip2}.
The three SigLIP2 variants use the same frozen model: zero-shot matching
compares query images with food names, gallery matching compares them with
one reference image per food, and fused matching averages both similarity
scores. To assess how language models use the retrieved evidence, we pair
SigLIP2 zero-shot matches with Llama-3.1-8B or Qwen3-8B readers using
Records, KAPING, zero-shot chain-of-thought (CoT), or
Think-on-Graph~\citep{baek2023kaping,kojima2022zeroshot,sun2024tog}.
Recommendation models learn from training respondents' intake and rank
SigLIP2-matched foods within rule-outcome groups. Text-only models provide
image-blind controls.
Model-specific objectives, checkpoint selection,
and representations of unseen respondents appear in
Appendix~\ref{sec:app-models}.

\paragraph{Inference and test protocol.}
We evaluate direct models zero-shot and count outputs that remain invalid
after normalization as errors (Appendix~\ref{sec:app-design-scoring}).
VQA scores average the listed image order and one seeded reordering.
To distinguish sampling uncertainty from variation across training runs,
we report clustered-bootstrap SE for single-run systems and three-seed SD
for trained models and the sampled Llama VQA and substitution controls.
Other language-model runs use greedy decoding. For seeded S2, we pool
Request and S1 totals across seeds and report the SD of seed-specific
ratios (Appendix~\ref{sec:app-stats}). We mark order-invariant inputs with
$^\circ$, privileged inputs with \priv{}, and native thinking mode with
(R); native thinking is otherwise disabled. Llama and Qwen abbreviate
Llama-3.1-8B and Qwen3-8B.
All configuration choices, including matcher selection, reasoning budgets, and recommender
training, were made on the validation split or on held-out training respondents and frozen
before test evaluation; the test split was used only for the final reported runs
(Appendix~\ref{sec:app-ledger}). Appendices~\ref{sec:app-models}, \ref{sec:app-retrained},
and~\ref{sec:app-t3} describe the evaluated configurations and task adaptations in more detail.

\Needspace{430pt}
\section{Task 1: \tvqa}
\label{sec:task-vqa}

\begin{wraptable}{r}{0.53\textwidth}
\centering\tabfont
\setlength{\tabcolsep}{3pt}
\begin{adjustbox}{max width=\linewidth}
\begin{tabular}{@{}lcc@{}}
\toprule
\textbf{System} & \textbf{Question} & \textbf{Pair} \\
\midrule
\rowcolor{groupgray}\multicolumn{3}{@{}l}{\textit{Text-only language models (image-blind controls)}} \\
Llama-3.1-8B$^{\circ}$ & 0.487\pmse{0.005} & 0.043\pmse{0.004} \\
Qwen3-8B$^{\circ}$ & 0.440\pmse{0.006} & 0.006\pmse{0.002} \\
\rowcolor{groupgray}\multicolumn{3}{@{}l}{\textit{Direct vision--language models}} \\
MiMo-VL-7B-RL & 0.835\pmse{0.012} & 0.696\pmse{0.023} \\
GLM-4.1V-9B (R) & 0.867\pmse{0.011} & 0.745\pmse{0.021} \\
InternVL3.5-8B & 0.621\pmse{0.010} & 0.268\pmse{0.019} \\
Qwen3-VL-2B & 0.584\pmse{0.008} & 0.219\pmse{0.015} \\
Qwen3-VL-4B & 0.680\pmse{0.011} & 0.428\pmse{0.021} \\
Qwen3-VL-8B & 0.726\pmse{0.011} & 0.501\pmse{0.021} \\
Qwen3-VL-8B (R) & 0.873\pmse{0.012} & 0.765\pmse{0.023} \\
\rowcolor{groupgray}\multicolumn{3}{@{}l}{\textit{Image matching + rules}} \\
DINOv2$^{\circ}$ & 0.722\pmse{0.024} & 0.599\pmse{0.034} \\
SigLIP2 zero-shot$^{\circ}$ & \textbf{0.922}\pmse{0.014} & \textbf{0.880}\pmse{0.021} \\
SigLIP2 gallery$^{\circ}$ & 0.684\pmse{0.023} & 0.520\pmse{0.034} \\
SigLIP2 fused$^{\circ}$ & 0.827\pmse{0.019} & 0.736\pmse{0.029} \\
Qwen3-VL-8B & 0.838\pmse{0.015} & 0.722\pmse{0.026} \\
\rowcolor{groupgray}\multicolumn{3}{@{}l}{\textit{Graph-RAG with SigLIP2 matches}} \\
Records (Llama) & 0.821\pmse{0.013} & 0.690\pmse{0.021} \\
Records (Qwen) & 0.796\pmse{0.013} & 0.652\pmse{0.021} \\
Records (Qwen, R) & \underline{0.921}\pmse{0.014} & 0.877\pmse{0.021} \\
CoT (Llama) & 0.836\pmse{0.011} & 0.698\pmse{0.020} \\
CoT (Qwen) & 0.917\pmse{0.013} & 0.868\pmse{0.021} \\
KAPING (Llama) & 0.839\pmse{0.013} & 0.712\pmse{0.020} \\
KAPING (Qwen) & 0.838\pmse{0.013} & 0.714\pmse{0.021} \\
KAPING (Qwen, R) & \textbf{0.922}\pmse{0.014} & \underline{0.878}\pmse{0.021} \\
ToG (Llama) & 0.867\pmse{0.013} & 0.771\pmse{0.020} \\
ToG (Qwen) & 0.884\pmse{0.013} & 0.799\pmse{0.021} \\
\bottomrule
\end{tabular}

\end{adjustbox}
\captionsetup{justification=raggedright,singlelinecheck=false}
\caption{We report VQA accuracy with bootstrap SE, except that Llama uses three-seed SD.
Bold and underlined values are the highest and second-highest scores.
}
\label{tab:t1-main}
\end{wraptable}

Answering a nutrient question about a food image requires selecting the
appropriate food record and using its values correctly. Recognizing a
broad category such as ``pasta'' does not establish which nutrient
profile applies. \tvqa{} therefore evaluates closed-set visual
food-to-record grounding followed by nutrient question answering.
Each instance supplies same-category candidate records, and the model
returns both its record assignment and its answer. The test set contains
1{,}758 nutrient-amount questions, 788 nutrient-comparison questions,
and 162 threshold questions. These require extracting a nutrient value,
comparing values across foods, or checking a nutrient limit,
respectively. Amount and threshold questions present one image,
while comparisons present two.

To make record selection consequential, we exclude questions whose
answers remain correct under an incorrect assignment. Paired instances
share the question and candidate nutrient values but depict different
foods requiring different answers. These pairs test whether models
adjust their answers to changes in visual evidence. We report question
accuracy over 2{,}708 questions and pair accuracy over 1{,}290 pairs.
Pair accuracy requires both answers to be correct, distinguishing
individual-question performance from success on both sides of a visual
contrast. Scoring uses exact matching, with a relative tolerance for
nutrient amounts (Appendix~\ref{sec:app-design-scoring}).

With SigLIP2 zero-shot matches, deterministic rules and KAPING (Qwen, R)
both achieve 0.922 question accuracy (Table~\ref{tab:t1-main}).
The rule pipeline achieves 0.880 pair accuracy, compared with 0.765
for the leading direct model, Qwen3-VL-8B (R). However, DINOv2 and
SigLIP2 gallery matching perform below the leading direct models
when using the same rules, indicating that matcher choice remains
consequential. Correct grounding does not ensure correct answers
from language-model readers either. With correct records, Records
(Qwen) answers only 0.559 of nutrient comparisons correctly
(Appendix~\ref{sec:app-why-t1}). These results identify two distinct
challenges: selecting the appropriate food record and correctly
using its nutrient values. Section~\ref{sec:analysis} examines
both sources of error.

\section{Task 2: \trec}
\label{sec:task-rec}

A food can satisfy a person's nutrient limits without matching their observed intake patterns.
\trec{} evaluates this distinction by ranking food options under explicit nutrient constraints and using a subsequent dietary recall to define an intake-derived relevance target.
Each request provides health conditions, per-serving
nutrient limits, first-recall intake, and eight option images arranged as
four same-category pairs, each containing one compliant and one
noncompliant food. Direct image inputs omit food names and nutrient
records, making food identification necessary.
Of the four compliant foods, two are relevant because their categories appear in the person's
second recorded recall.
The 406 requests form 203 respondent pairs over 58 option-image sets. Each pair shares conditions, limits, and food options but differs in intake history and relevance labels, so identical selections cannot satisfy both respondents. However, independently shuffled option positions prevent fully isolating intake-history effects from presentation order.

We evaluate the top two ranked options and emphasize \emph{Request
accuracy}, which requires both selections to satisfy every nutrient limit
and receive positive relevance labels. Two component metrics support
diagnosis: S1 measures the proportion of requests with both selections
compliant, while S2 measures relevance conditional on satisfying S1.
Request accuracy equals S1 multiplied by S2 within a run. Since systems
may satisfy S1 on different request subsets, conditional S2 alone cannot
establish their relative end-to-end performance. For seeded S2, we pool
the numerator and denominator across seeds. Chance performance is 0.214
for S1 and 0.167 for S2. Here, \emph{safety} denotes nutrient-rule
compliance, not clinical safety.
A positive relevance label indicates that the food's category appears in the respondent's later recorded intake, while a negative label indicates its absence. Since absence does not imply dislike or rejection, the task measures alignment with recorded intake rather than explicit food preferences.

\begin{table}[t]
\centering\fontsize{9}{10.6}\selectfont\let\scriptsize\uncertaintyfont
\setlength{\tabcolsep}{3pt}\renewcommand{\arraystretch}{1.0}
\begin{minipage}[t]{0.49\linewidth}
\vspace{0pt}
\centering
\begin{adjustbox}{max width=\linewidth}
\begin{tabular}{@{}lccc@{}}
\toprule
\textbf{System} & \textbf{S1} & \textbf{S2} & \textbf{Request} \\
\midrule
\rowcolor{groupgray}\multicolumn{4}{@{}l}{\textit{Reference orders after matching and rules}} \\
Random order & 0.559\pmse{0.043} & 0.170\pmse{0.003} & 0.095\pmse{0.007} \\
Popularity order & 0.627\pmse{0.067} & 0.190\pmse{0.018} & 0.119\pmse{0.017} \\
\rowcolor{groupgray}\multicolumn{4}{@{}l}{\textit{Direct vision--language models}} \\
MiMo-VL-7B-RL & 0.328\pmse{0.043} & 0.173\pmse{0.029} & 0.057\pmse{0.012} \\
GLM-4.1V-9B (R) & 0.303\pmse{0.037} & 0.138\pmse{0.031} & 0.042\pmse{0.011} \\
InternVL3.5-8B & 0.175\pmse{0.028} & 0.127\pmse{0.046} & 0.022\pmse{0.008} \\
Qwen3-VL-2B & 0.202\pmse{0.016} & 0.122\pmse{0.037} & 0.025\pmse{0.008} \\
Qwen3-VL-4B & 0.217\pmse{0.031} & 0.148\pmse{0.039} & 0.032\pmse{0.008} \\
Qwen3-VL-8B & 0.155\pmse{0.033} & 0.159\pmse{0.053} & 0.025\pmse{0.008} \\
\rowcolor{groupgray}\multicolumn{4}{@{}l}{\textit{Text-only language models (image-blind controls)}} \\
Llama-3.1-8B & 0.214\pmse{0.016} & 0.172\pmse{0.039} & 0.037\pmse{0.008} \\
Qwen3-8B & 0.229\pmse{0.017} & 0.183\pmse{0.043} & 0.042\pmse{0.010} \\
\rowcolor{groupgray}\multicolumn{4}{@{}l}{\textit{Graph-RAG with SigLIP2 matches}} \\
Records (Llama) & 0.369\pmse{0.041} & 0.147\pmse{0.028} & 0.054\pmse{0.013} \\
Records (Qwen) & 0.249\pmse{0.030} & 0.178\pmse{0.034} & 0.044\pmse{0.010} \\
Records (Qwen, R) & 0.495\pmse{0.049} & 0.189\pmse{0.024} & 0.094\pmse{0.015} \\
CoT (Llama) & 0.424\pmse{0.034} & 0.157\pmse{0.026} & 0.067\pmse{0.012} \\
CoT (Qwen) & 0.519\pmse{0.046} & 0.213\pmse{0.023} & 0.111\pmse{0.015} \\
KAPING (Llama) & 0.360\pmse{0.043} & 0.185\pmse{0.027} & 0.067\pmse{0.012} \\
KAPING (Qwen) & 0.266\pmse{0.029} & 0.176\pmse{0.030} & 0.047\pmse{0.009} \\
KAPING (Qwen, R) & 0.340\pmse{0.041} & 0.167\pmse{0.028} & 0.057\pmse{0.012} \\
\bottomrule
\end{tabular}

\end{adjustbox}
\end{minipage}\hfill
\begin{minipage}[t]{0.49\linewidth}
\vspace{0pt}
\centering
\begin{adjustbox}{max width=\linewidth}
\begin{tabular}{@{}lccc@{}}
\toprule
\textbf{System} & \textbf{S1} & \textbf{S2} & \textbf{Request} \\
\midrule
\rowcolor{groupgray}\multicolumn{4}{@{}l}{\textit{Graph-RAG with SigLIP2 matches (continued)}} \\
ToG (Llama) & 0.372\pmse{0.041} & 0.159\pmse{0.027} & 0.059\pmse{0.013} \\
ToG (Qwen) & 0.340\pmse{0.039} & 0.159\pmse{0.026} & 0.054\pmse{0.011} \\
\rowcolor{groupgray}\multicolumn{4}{@{}l}{\textit{Image matching + rules}} \\
DINOv2 & 0.286\pmse{0.046} & 0.172\pmse{0.025} & 0.049\pmse{0.011} \\
SigLIP2 zero-shot & 0.592\pmse{0.057} & 0.189\pmse{0.023} & 0.112\pmse{0.018} \\
SigLIP2 gallery & 0.268\pmse{0.045} & 0.197\pmse{0.036} & 0.053\pmse{0.011} \\
SigLIP2 fused & 0.365\pmse{0.045} & \underline{0.226}\pmse{0.026} & 0.083\pmse{0.012} \\
Qwen3-VL-8B & 0.366\pmse{0.056} & 0.165\pmse{0.030} & 0.060\pmse{0.015} \\
\rowcolor{groupgray}\multicolumn{4}{@{}l}{\textit{Trained rankers after matching and rules}} \\
MF-BPR & 0.571\pmse{0.024} & 0.205\pmse{0.034} & 0.117\pmse{0.018} \\
GCN & 0.539\pmse{0.014} & 0.207\pmse{0.017} & 0.112\pmse{0.011} \\
GraphSAGE & 0.560\pmse{0.030} & 0.213\pmse{0.008} & 0.119\pmse{0.010} \\
GAT & 0.544\pmse{0.033} & 0.172\pmse{0.021} & 0.094\pmse{0.015} \\
NGCF & 0.557\pmse{0.023} & 0.205\pmse{0.008} & 0.114\pmse{0.002} \\
LightGCN & 0.576\pmse{0.024} & 0.223\pmse{0.019} & \underline{0.128}\pmse{0.016} \\
SGL & 0.576\pmse{0.025} & \textbf{0.229}\pmse{0.019} & \textbf{0.132}\pmse{0.017} \\
SimGCL & 0.580\pmse{0.024} & 0.220\pmse{0.011} & 0.128\pmse{0.010} \\
RecipeRec & 0.527\pmse{0.019} & 0.202\pmse{0.007} & 0.106\pmse{0.005} \\
LightGCL & \textbf{0.623}\pmse{0.021} & 0.205\pmse{0.016} & 0.128\pmse{0.013} \\
HFRS-DA & \underline{0.600}\pmse{0.039} & 0.189\pmse{0.017} & 0.113\pmse{0.017} \\
MOPI-HFRS & \textbf{0.623}\pmse{0.007} & 0.187\pmse{0.006} & 0.117\pmse{0.004} \\
\bottomrule
\end{tabular}

\end{adjustbox}
\end{minipage}
\caption{We report recommendation compliance (S1), conditional relevance (S2), and complete-request
accuracy. Matching pipelines and rankers share the rule filter but differ in grounding or ranking.
Spreads are clustered-bootstrap SE for single runs and reference orders, and three-seed SD for
trained rankers; S2 pools seeds. Bold and underlined values mark the highest and second-highest
scores.}
\label{tab:t2-main}
\end{table}

Matching pipelines prioritize foods predicted to satisfy the limits and
use a naive-Bayes estimator based on first-recall categories to rank
foods within each rule-outcome group. The true-food reference uses the
same procedure with correct identities.
Under the same rule filter and ranking procedure, SigLIP2 achieves S1 of 0.592, compared with 0.286 for DINOv2 (Table~\ref{tab:t2-main}), indicating how the choice of image-to-record matcher affects downstream rule compliance.
However, higher compliance does not necessarily yield better intake
alignment: MOPI-HFRS achieves higher S1 than SGL (0.623 versus 0.576)
but lower Request accuracy (0.117 versus 0.132). With the shared SigLIP2
filter, SGL also exceeds random ordering in Request accuracy
(0.132 versus 0.095); its S2 of 0.229 describes relevance only among
compliant requests. These differences may reflect intake prediction or
the ordering of incorrectly filtered foods. Section~\ref{sec:analysis}
examines these explanations, with additional controls in
Appendix~\ref{sec:app-more-results}.
Complete-request accuracy remains low across the evaluated systems, indicating that satisfying nutrient constraints and matching intake-derived relevance labels jointly remains challenging. These results also illustrate why compliance alone is insufficient to characterize performance on the constructed recommendation task.

\Needspace{420pt}
\section{Task 3: \tsub}
\label{sec:task-sub}

\begin{wraptable}{r}{0.53\textwidth}
\centering\tabfont
\setlength{\tabcolsep}{3pt}
\begin{minipage}[t]{\linewidth}
\vspace{0pt}\centering
\begin{adjustbox}{max width=\linewidth}
\begin{tabular}{@{}lcc@{}}
\toprule
\textbf{System} & \textbf{Pair} & \textbf{Instance} \\
\midrule
\rowcolor{groupgray}\multicolumn{3}{@{}l}{\textit{Graph-RAG with SigLIP2 matches}} \\
Records (Llama) & 0.013\pmse{0.005} & 0.221\pmse{0.011} \\
Records (Qwen) & 0.005\pmse{0.003} & 0.180\pmse{0.010} \\
Records (Qwen, R) & 0.861\pmse{0.014} & 0.928\pmse{0.007} \\
KAPING (Llama) & 0.020\pmse{0.006} & 0.254\pmse{0.011} \\
KAPING (Qwen) & 0.008\pmse{0.004} & 0.170\pmse{0.010} \\
KAPING (Qwen, R) & 0.793\pmse{0.016} & 0.890\pmse{0.009} \\
CoT (Llama, 2{,}048 tokens) & 0.110\pmse{0.013} & 0.341\pmse{0.013} \\
CoT (Qwen, 8{,}192 tokens) & 0.582\pmse{0.020} & 0.761\pmse{0.013} \\
ToG (Llama) & 0.011\pmse{0.004} & 0.191\pmse{0.010} \\
ToG (Qwen) & 0.005\pmse{0.003} & 0.148\pmse{0.009} \\
\rowcolor{groupgray}\multicolumn{3}{@{}l}{\textit{Direct vision--language models}} \\
MiMo-VL-7B-RL & 0.779\pmse{0.017} & 0.884\pmse{0.009} \\
GLM-4.1V-9B (R) & 0.751\pmse{0.018} & 0.866\pmse{0.010} \\
Qwen3-VL-8B (R) & 0.644\pmse{0.020} & 0.795\pmse{0.012} \\
\rowcolor{groupgray}\multicolumn{3}{@{}l}{\textit{Image matching + rules}} \\
DINOv2 & 0.639\pmse{0.019} & 0.798\pmse{0.012} \\
SigLIP2 zero-shot & \textbf{0.936}\pmse{0.010} & \textbf{0.967}\pmse{0.005} \\
SigLIP2 fused & 0.800\pmse{0.016} & 0.899\pmse{0.008} \\
\rowcolor{groupgray}\multicolumn{3}{@{}l}{\textit{Supervised nutrient heads with SigLIP2 matches}} \\
Nutrient values only & \underline{0.923}\pmse{0.006} & \underline{0.961}\pmse{0.003} \\
TransE & 0.909\pmse{0.003} & 0.953\pmse{0.002} \\
ComplEx & 0.887\pmse{0.014} & 0.942\pmse{0.006} \\
R-GCN & 0.848\pmse{0.020} & 0.919\pmse{0.011} \\
RotatE & 0.844\pmse{0.003} & 0.920\pmse{0.001} \\
\bottomrule
\end{tabular}
\end{adjustbox}
\end{minipage}

\captionsetup{justification=raggedright,singlelinecheck=false}
\caption{We report substitution accuracy with clustered-bootstrap SE, except that supervised
heads use SD over 3 seeds. Bold and underlined values mark the highest and second-highest scores.}
\label{tab:t3-main}
\end{wraptable}

A replacement food must satisfy both a nutrient-reduction goal and the remaining stated nutrient limits.
\tsub{} evaluates whether a system identifies the source food from its image, compares candidate nutrient records with the source, and selects a replacement that satisfies the complete rule.
Each instance provides a source image, records for two possible sources,
and eight same-family candidates with nutrient records. A valid
replacement must reduce the target nutrient per serving by at least
20\%, increase no other constrained nutrient by more than 25\%, and
satisfy every personal limit. Some distractors meet the reduction goal
but violate another constraint. The system must select the unique valid replacement
or return none.

Paired instances share the text and candidate pool but depict different
sources requiring different answers. A deterministic image-blind response
cannot answer both correctly. We report instance accuracy over 1{,}220
decisions and pair accuracy over 610 pairs, requiring both answers in
a pair to be correct. To limit serving-size effects, we retain candidates
whose labels agree when source-relative constraints use either per-serving
or per-100-g values; personal limits remain per serving. Additional
health-condition limits change the correct answer in only 2.2\% of
instances. The task thus primarily evaluates source-dependent nutrient
comparison and complete constraint checking, with limited coverage of
decisions that depend on condition-specific limits
(Appendix~\ref{sec:app-t3}).

SigLIP2 zero-shot matching with rules achieves 0.936 pair accuracy,
compared with 0.861 for Records (Qwen, R), 0.793 for KAPING (Qwen, R),
and 0.779 for the leading direct model, MiMo-VL-7B-RL
(Table~\ref{tab:t3-main}). Replacing SigLIP2 with DINOv2 reduces pair
accuracy to 0.639.
This difference shows that source-record matching substantially affects the evaluated substitution pipeline.
With SigLIP2 matches, the supervised nutrient-values-only head achieves
0.923, while heads incorporating graph representations achieve
0.844--0.909. These comparisons involve different training procedures
and do not isolate the effect of graph structure
(Appendix~\ref{sec:app-t3-training-audit}). Correct grounding is also
insufficient for some readers. With true source identities, CoT (Llama)
achieves only 0.085 pair accuracy and continues to violate additional
constraints and miss cases requiring \emph{none}. These errors highlight
complete rule application and rejection as challenges even when the
source record is correct (Section~\ref{sec:analysis};
Table~\ref{tab:t3-full}).

\section{Analysis of Benchmark Challenges}
\label{sec:analysis}

We examine grounding, rule-application, and intake-relevance errors using linked records and input controls (Appendix~\ref{sec:app-why}).

\Needspace{225pt}
\begin{wrapfigure}{r}{0.48\textwidth}
\centering
\includegraphics[width=\linewidth]{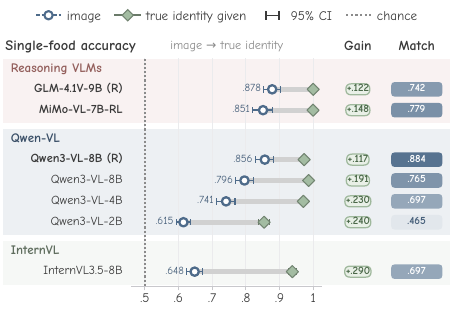}
\captionsetup{justification=raggedright,singlelinecheck=false}
\caption{Supplying true identities improves single-food VQA accuracy. Match denotes exact-record
accuracy (Table~\ref{tab:t1-full}). Bold names identify the three highest-scoring direct models
in Table~\ref{tab:t1-main}.}
\label{fig:t1-single-food}
\end{wrapfigure}

\paragraph{VQA separates grounding from comparison reading.}
To determine whether better identification alone can improve VQA, we
examine errors after correct food matching. Every SigLIP2-and-rules error
involves an incorrect match (Table~\ref{tab:app-error-types}), but readers
can still fail with correct records: native-thinking Graph-RAG readers
achieve 0.999 accuracy, whereas Records (Qwen) answers only 0.559 of
comparisons correctly (Appendix~\ref{sec:app-why-t1}). Supplying true
identities also improves direct-model accuracy by $+$0.117--$+$0.290
on 1{,}920 single-food questions (Figure~\ref{fig:t1-single-food}),
although output completion changes too
(Appendix~\ref{sec:app-control-output}). Better grounding can therefore
address the rule pipeline's errors, while readers that fail with correct
records also need more reliable use of nutrient values.

\finishwrap
\Needspace{315pt}
\begin{wrapfigure}{r}{0.48\textwidth}
\centering
\includegraphics[width=\linewidth]{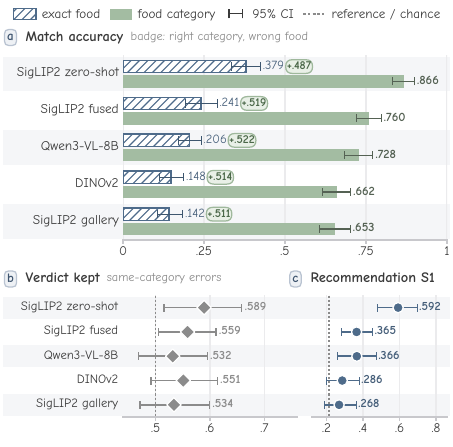}
\captionsetup{justification=raggedright,singlelinecheck=false}
\caption{Category recognition does not reliably preserve nutrient-rule outcomes. Panels show
(a) Record vs. category matching
(b) Rule preservation
(c) Compliance (S1) in our within-category contrasts.}
\label{fig:analysis-precision}
\end{wrapfigure}

\paragraph{Recommendation requires correct rule outcomes and person-specific ranking.}
MOPI-HFRS and HFRS-DA achieve higher compliance than LightGCN but lower
Request accuracy, prompting us to examine ranking separately from
filtering. Grounding can affect this comparison: incorrect same-category
matches preserve rule outcomes only 0.532--0.589 of the time, close to
the 0.5 reference (Figure~\ref{fig:analysis-precision}). With true identities
and selection restricted to four compliant foods, without retraining,
LightGCN and SGL recover both relevant foods on 0.191 and 0.206 of
requests, versus 0.172 for MOPI-HFRS and 0.171 for HFRS-DA.
In the paired relevance evaluation, MOPI-HFRS selects the same top-two foods for both respondents in 67.5\% of pairs, compared with 35.8\% for HFRS-DA and 22.8\% for LightGCN (Appendix~\ref{sec:app-why-t2}).
These repeated selections cannot satisfy both respondents' contrasting relevance labels, although they do not establish an architectural cause. Repeating first-recall categories with true foods
achieves Request accuracy of 0.234
(Appendix~\ref{sec:app-preference}), showing that history remains
informative. The challenge is to use it to distinguish relevant choices,
while recognizing that later intake provides incomplete preference feedback.

\finishwrap
\paragraph{Substitution requires source grounding and complete rule application.}
Supplying true source identities raises MiMo-VL-7B-RL pair accuracy from 0.779 to 0.962, but CoT (Llama) reaches only 0.085 (Table~\ref{tab:t3-full}). For CoT (Llama), 909 of 1{,}032 selections meet the reduction goal, but only 419 satisfy the other-nutrient clause. It also correctly returns \emph{none} on only 86 of 440 instances without valid replacements, despite low invalid-output and truncation rates (Appendix~\ref{sec:app-why-t3}). Thus, correct grounding does not ensure complete constraint checking or rejection. Differences in completion rates, token budgets, and answer prefills prevent attributing performance gaps to thinking mode alone.

We report further qualitative analysis about the dataset and benchmark tasks in Appendix~\ref{sec:app-cases}.

\finishwrap
\section{Related Work}
\label{sec:related}

Food-image research evaluates recognition, nutrient estimation, and dietary knowledge. Nutrition5k links food images with measured ingredient masses and nutrients~\citep{thames2021nutrition5k}, while FoodieQA and WorldCuisines assess cultural food understanding~\citep{li2024foodieqa,winata2025worldcuisines}. DietAI24 retrieves FNDDS records for nutrient estimation~\citep{yan2025dietai24}, and FAM-Bench evaluates condition-specific dish suitability and ranking~\citep{mao2026fambench}. These benchmarks address complementary aspects of food understanding, but differ in how visual identification is connected to subsequent dietary decisions.

Structured nutrition resources support constrained queries and recommendation~\citep{haussmann2019foodkg,chen2021pfoodreq,mohbat2025kerl}. NGQA and GLEN-Bench extend this direction to survey-grounded nutritional reasoning~\citep{zhang2024nutrition,glenbench2026}, while GISMo uses recipe context and graph relations for ingredient substitution~\citep{fatemi2023gismo}. However, these reasoning tasks generally assume known food or ingredient identities rather than requiring visual food-to-record grounding. Consequently, they do not directly examine how incorrect record assignments affect downstream nutrient-rule decisions.

Multimodal benchmarks also study visual entity recognition and knowledge-intensive question answering~\citep{hu2023oven,chen2023infoseek,mensink2023encyclopedic}. Related work connects images with patient records~\citep{bae2023ehrxqa} or enriches entity representations with multimodal information~\citep{zhang2024native,zhang2021lattice}. Building on these directions, \bench{} links independently collected food images with nutrient records and respondent intake histories within a shared evaluation. Its paired tasks examine how visual identity and respondent context affect decisions, while input controls help distinguish record-selection errors from subsequent evidence-use errors.

\section{Conclusion}
\label{sec:conclusion}

\bench{} connects food images, nutrient records, and respondent intake
histories to evaluate dietary decisions under explicit constraints.
We examine how visual identity, respondent context, and evidence affect decisions.
The results identify three complementary challenges. Grounding errors
lead systems to select the wrong nutrient records. Rule-application
errors persist for some readers even when the correct records are
provided.
Intake-relevance errors occur when compliant selections fail to match the categories observed in a respondent's later dietary recall.
These findings show why record selection, complete nutrient-rule application, and intake-derived ranking should be evaluated separately rather than summarized by a single measure of dietary decision quality.
The catalog and results dashboard provide access to the records and evaluation conventions for inspection.

\section*{Limitations}

\bench{} evaluates constructed diagnostic cases rather than clinical dietary suitability or health outcomes. Its results depend on image-to-food associations that have not been fully human-validated. The audit covers only the initial corpus, and some assigned records do not describe the entire depicted meal. Model comparisons are also descriptive because evaluated configurations differ in evidence access, adaptation, and inference settings. These differences limit conclusions about architectural superiority. Future work could involve nutrition experts, more representative food images, and real-world evaluations. Longer-term intake records and explicit feedback could also improve the assessment of dietary preferences and personalized recommendations.

\section*{Ethics Statement}
UMAMI is derived from public-use, de-identified NHANES records and independently collected food
images. Task instances use tokenized respondent identifiers, and their use is subject to the
NCHS data use terms~\citep{nchs2024dua}, which prohibit attempts to identify respondents.
Image licenses, release terms, and the removal of eight images flagged for unsafe content are
described in Appendices~\ref{sec:app-audit} and~\ref{sec:app-datasheet}. The dataset is
intended for research rather than clinical dietary advice. Three project members conducted the
image audit with informed consent, and AI assistants supported
code development and writing.

\FloatBarrier
\bibliographystyle{plainnat}
\bibliography{refs}
\clearpage
\appendix
\raggedbottom

\section{Dataset Construction and Quality}
\label{sec:app-resource}

\begin{figure}[t]
\centering
\includegraphics[width=\linewidth]{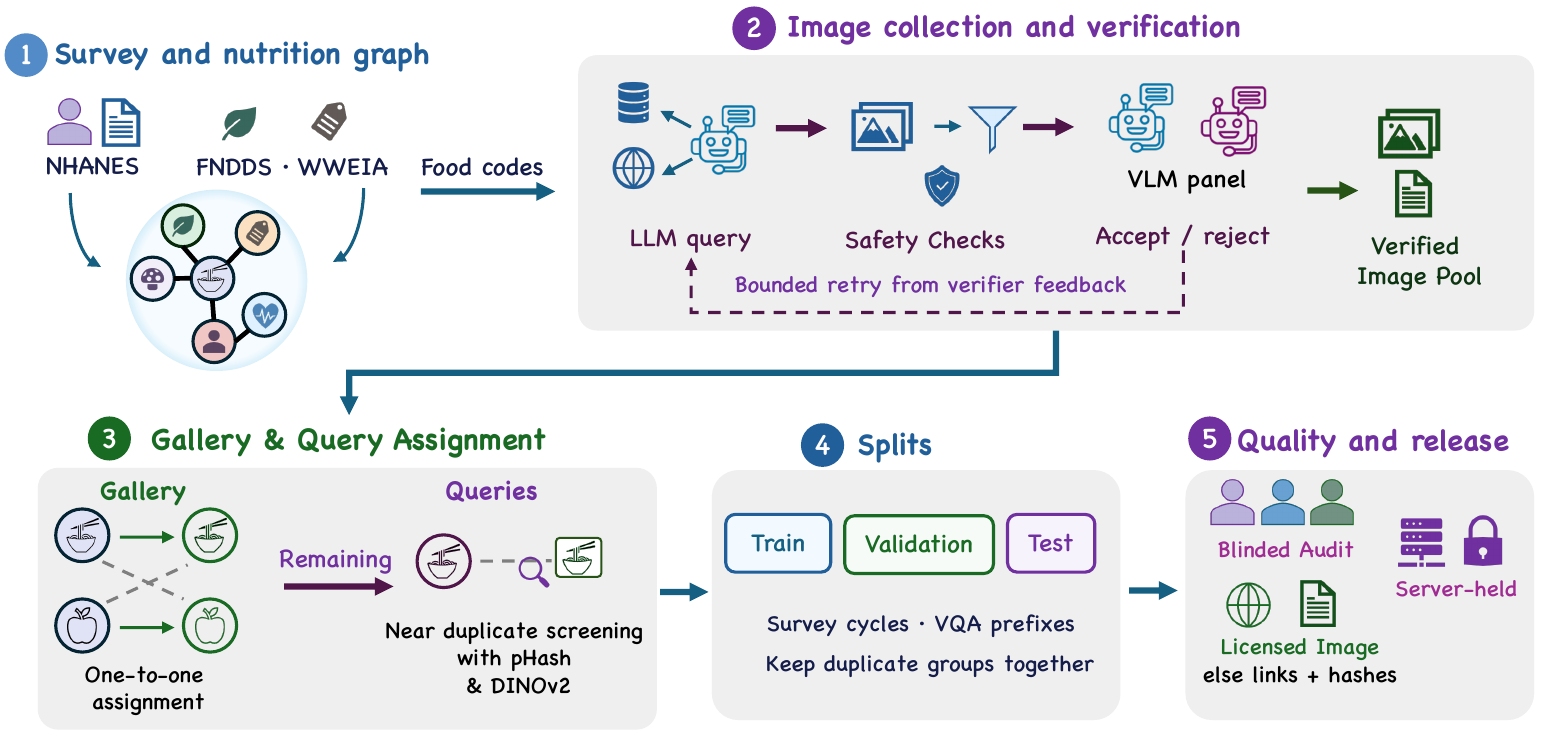}
\caption{Construction links survey records and independently collected images through food codes.
Gallery and query roles separate retrieval evidence from task inputs.}
\label{fig:pipeline}
\end{figure}

Food codes link independently collected images to composition and intake records
(Figure~\ref{fig:pipeline}); the images do not document respondents' meals. We partition the resource and define graph views to control evidence access.
The collection pipeline establishes candidate image-to-food associations, while a stratified human audit assesses collection acceptance in the initial corpus. The audit does not independently validate the test-only extension or the generated benchmark answers.
Table~\ref{tab:dataset-stats} summarizes resource and task sizes. The audit assesses image
acceptance, not benchmark-answer validity.

\paragraph{Respondent and image splits.} We partition respondents by survey release so that
recommendation can use a held-out person's intake history without training on their evaluation
records: the training
releases (2003--2014, 54{,}572 respondents) provide all behavioral signals and recommender training
data; the 2015--2016 release (8{,}505) provides validation respondents for \trec{}; and the
2017--March 2020 release (12{,}634) provides test respondents. These counts exclude 146 possible
duplicates retained in the upstream resource total. The first and later recalls cover
nonconsecutive days, so the prediction target is a later intake occasion rather than an
adjacent-day transition.

Respondent separation does not prevent reuse of visual evidence. We therefore assign duplicate
image groups, rather than individual files, to tasks.
Hashing these groups assigns 30\% to \tvqa{} and the remainder to \trec{}, keeping an image and its
near-duplicates together. Within \tvqa{}, hashing each food's six-digit code prefix assigns 40\%
of prefixes to validation, so foods sharing a prefix remain in the same split. Images used in an
earlier benchmark build are restricted to validation; newly collected images extend the test sets.
File hashes, perceptual hashes, and DINOv2 similarity identify no \tvqa{} validation image in
another split or task. We apply the same overlap criteria when constructing \trec{}, while
\tsub{} uses a separate image pool and overlap check
(Appendices~\ref{sec:app-checks} and~\ref{sec:app-t3}).

\paragraph{Graph views.} We separate static food composition from behavioral evidence to
restrict learned intake associations to training respondents.
The static component contains foods, WWEIA categories, ingredients, and the
gallery, together with FNDDS 2017--2018 nutrient values per 100\,g and values for the ``quantity not
specified'' serving. Images have only image-to-food edges. We use FNDDS 2017--2018 throughout,
without replacing missing profiles or servings with imputations or per-release medians. Each task
requires the nutrient values used in its questions or rules; recommendation and substitution also
require the database serving. Using one nutrient version and explicit measurement bases makes
the rule targets reproducible across survey releases; it does not impose a historical knowledge cutoff.
The nutrient profiles were published in October 2020; category names use the October 2024
WWEIA dictionary, with inherited GLEN names for categories absent from that dictionary.

The behavioral component contains co-consumption, popularity, and person--food edges derived
exclusively from training releases. Validation and test respondents are absent from this graph.
We build co-consumption counts in the recorded environment (NumPy 2.2.6 and SciPy 1.15.3)
and check an independent reconstruction, including symmetry and a zero diagonal.
\trec{} accesses both components and the person's first-recall foods. The public inputs for
\tvqa{} contain the two candidate food records; \tsub{} supplies records for two possible source
foods and eight candidates. VQA Graph-RAG can additionally retrieve food neighborhoods from
the permitted graph, including training co-consumption relations. Its linker is restricted to the
two candidate foods, but its retrieved evidence can extend beyond their supplied records.

\paragraph{Health conditions.} We derive four condition indicators for adults aged 18 or more
from examination, laboratory, and questionnaire records. These indicators assign the applicable
nutrient constraints, so the same food can face different rules for different people. Hypertension is defined by mean measured blood
pressure of at least 130/80\,mm\,Hg or current blood-pressure medication. Diabetes is defined by a
reported diagnosis or HbA1c of at least 6.5\%. High cholesterol is defined by total cholesterol of at
least 240\,mg/dL or current cholesterol medication. Obesity is defined by a body-mass index of at least
30. Missing measurements produce an unknown status, not a negative label.
Table~\ref{tab:app-rule-fail-dev} reports the corresponding test-set condition counts.

\paragraph{Collecting and verifying images.} We collect candidates from Recipe1M, Openverse,
Wikimedia Commons, Open Food Facts, and DuckDuckGo web search, then screen whether each image
supports its assigned food description.
The final collection pass combines checks for food identity,
single-subject presentation, and unsafe content. Qwen2.5-VL-7B must match the food
description with confidence of at least 70 and accept the single-subject presentation. Idefics3-8B
must independently accept the depicted food. CLIP ViT-L/14 must not reject the image as unrelated
to its food name, and the Falconsai safety classifier must assign an unsafe-content score below
0.5. Qwen2.5-VL-7B and Idefics3-8B form the final verification panel, and every test-only extension
image meets this rule.

The initial corpus also contains images accepted under earlier panels:
Qwen2.5-VL-7B alone, or a two-of-three vote of Qwen2.5-VL-7B, LLaVA-1.5-7B, and CLIP ViT-L/14. Of the
10{,}179 corpus query images, 4{,}938 have no recorded Idefics3-8B vote, and 732 have no recorded
Qwen2.5-VL-7B acceptance: 683 were accepted by LLaVA-1.5-7B and CLIP ViT-L/14 against Qwen2.5-VL-7B,
and 49 by Idefics3-8B against Qwen2.5-VL-7B. The initial corpus therefore does not share a uniform
acceptance panel with the extension. We exclude these vision--language verifiers and CLIP
from evaluation because they participated in image selection (Appendix~\ref{sec:app-ledger}).

\paragraph{Gallery and query assignment.} We use minimum-cost one-to-one assignment to select
a distinct reference image file for each of 8{,}821 foods. Gallery identifiers and file hashes are distinct,
but visual near-duplicates remain across food codes: 27 pairs have perceptual-hash distance at most 2,
including 22 with identical perceptual hashes; 26 have DINOv2 similarity of at least 0.98.
The assignment therefore prevents file reuse without guaranteeing visually distinct record evidence.
To evaluate retrieval on separate inputs, we draw query candidates from the remaining verified
images and screen them against the gallery. Each candidate must be associated with exactly one food code
and have a shorter side of at least 224 pixels. It must have no perceptual-hash
duplicate in the gallery and a DINOv2 cosine similarity below 0.98 to every gallery image.
The resulting 10{,}179 query images cover 6{,}000 foods. Of these images, 9{,}202 come
from a different source than their food's gallery image.

\paragraph{Test-only query images.} After fixing the validation sets, we collect additional
query images for the test extension. We exclude
images linked to multiple foods, corpus
duplicates, within-set near-duplicates, and images that fail the DINOv2 or size thresholds, as well
as foods without gallery images.
The extension contains 8{,}428 query images covering 2{,}606 foods, none
of which belongs to the \tvqa{} validation split. Most of them (8{,}065) come from DuckDuckGo web
search. They were accepted in a later collection pass that required Qwen2.5-VL-7B and Idefics3-8B to
agree on the food, no CLIP rejection, and a single clean subject; every accepted extension image has a
Qwen2.5-VL-7B confidence of at least 90. These stricter automated criteria do not establish human
validation: the image audit does not cover the extension.

\subsection{Human Audit}
\label{sec:app-audit}

The audit includes both retained and rejected images to assess disagreement with the collection
pipeline. We sample 300 accepted images, 109 rejected candidates, and 91 candidates
that the pipeline did not assess. Accepted images with low confidence or limited metadata are
oversampled; population weights account for this stratification. Three project members independently
assess all 500 items, blinded to the source pool and verification panel's decision.

Annotators assess food identity, presentation, visual prominence, apparent portion, acceptability,
and unsafe content. Their population-weighted false-acceptance rates are 7.9\%, 10.2\% and
11.7\%, compared with unweighted rates of 13.0\%, 15.5\% and 17.0\%.
Acceptance agreement is 82.2\%, with mean pairwise Cohen's $\kappa=0.40$. These estimates characterize
collection acceptance, not human task performance or independent validation of benchmark answers.
Portion judgments further bound what the images establish: the annotators could not determine
portion size for 93.0\%, 91.2\% and 86.6\% of items, respectively. Database-defined servings therefore
cannot be interpreted as estimates of depicted portions.
Of the 109 rejected candidates, the three annotators would accept 96, 90 and 89, respectively.
These unweighted sample counts show disagreement with automated rejection, not a population false-rejection
estimate; their acceptability judgments need not enforce the panel's exact-record criteria.

The audit also flagged 8 images for unsafe content despite automated screening. None was a task
query image; 7 were gallery references. We removed all 8 images from the released corpus
and replaced the 7 gallery references with other verified images of the same foods.
The three annotators are project members who
participated voluntarily as part of their research roles.
No personal information about annotators is collected.

\section{Task Construction and Input Conditions}
\label{sec:app-taskdetail}

\subsection{Question and Request Construction}

Each VQA question supplies same-category records in shuffled order and asks about a nutrient
threshold, amount, or comparison. We require all needed nutrient values to be present and every incorrect image-to-record
assignment to change the answer. Category recognition alone therefore cannot determine the correct
response. We balance threshold answers and comparison directions and fix paraphrases before
evaluation. The test set contains 162 threshold, 1{,}758 amount, and 788 comparison questions over
16 nutrients.

We pair respondents who share conditions, sex, age band, and the option-image set but differ
in intake history. Their relevant options form disjoint subsets of the foods satisfying the shared
rules, so compliance does not identify the relevant choice for both people. Option positions are
shuffled independently; changes across a pair consequently cannot be attributed to history alone.

To limit popularity and presentation shortcuts, all options in a request come from the same
training-popularity stratum. Each option image is relevant in half of the requests in which it
passes, preventing its marginal frequency from determining relevance. Pairing and balancing retain
406 of 4{,}731 eligible respondents, forming an intentionally contrasting challenge subset rather
than a representative survey sample. We exclude water, coffee, tea, and diet-drink categories from
candidate options and both recall histories. Consumption in the later recall supplies incomplete
implicit feedback: an unseen category is not proof of dislike.

\subsection{Nutrient Rules}
\label{sec:app-design-rules}

\paragraph{Per-100\,g rules for \tvqa{}.} A versioned registry specifies each operation,
nutrient field, threshold, unit, measurement basis,
source, and status.
A threshold question classifies a food as ``high'' in a nutrient when its value reaches 20\%
of the FDA Daily Value from the 2016 labeling rule~\cite{fda2016dailyvalue}.
The FDA's consumer guide states that ``20\% DV or more of a nutrient per
serving is considered
high''.\footnote{\def\UrlBreaks{\do\/\do\-\do\.}\url{https://www.fda.gov/food/nutrition-facts-label/how-understand-and-use-nutrition-facts-label}}
We adapt this threshold to the per-100\,g values stated in our questions rather than use it to
assess the depicted portion. The registry covers sodium, saturated fat, cholesterol, carbohydrate,
fiber, protein, calcium, iron, phosphorus, potassium, folic acid, and vitamins B12, C, and D;
test threshold questions use all of these except folic acid. A \emph{validated} registry
status documents the FDA threshold but does not establish human validation of this adaptation or
its labels. We exclude rules without a corresponding Daily Value from threshold questions and
retain their \emph{draft} status in the registry. Energy has no Daily Value. The FDA values for
added sugars and folate do not correspond to the registry's total-sugars and folic-acid fields.

\paragraph{Per-serving rules for \trec{} and \tsub{}.} We use each food's FNDDS 2017--2018
``quantity not specified'' serving and apply one rule per condition; substitution uses the same
thresholds as personal limits; neither task estimates the depicted portion. The hypertension rule rejects options
with at least 460\,mg of sodium, the
high-cholesterol rule rejects those with at least 4\,g of saturated fat, and the diabetes rule rejects
those with at least 55\,g of carbohydrate. These thresholds correspond to 20\% of the respective
Daily Values. The obesity rule rejects options with at least 400\,kcal, or 20\% of the
2{,}000\,kcal label reference. This energy threshold is a benchmark convention rather than a Daily Value.

Recommendation assigns an \emph{unknown} outcome near a threshold to avoid deciding compliance
from a small numerical difference: a value is unknown when it falls
within 10\% of the rule's threshold. We retain an option only when its overall label is known:
it must pass every rule or fail at least one, even if another outcome is unknown. A definite failure
therefore suffices to reject an option without resolving all its other rule outcomes. These labels
measure nutrient-rule compliance, not clinical suitability. Reduced sodium intake
can lower blood pressure~\cite{sacks2001dash}, which supports the sodium rule's direction but
does not validate its per-food threshold.

Table~\ref{tab:app-rule-fail-dev} reports how often each rule applies across 406 test requests,
14 condition sets, and 320 option foods. It counts respondents with each condition and foods with
passing, failing, or unknown outcomes. These counts describe the constraint distribution, not the
clinical suitability of the options.

\begin{table}[t]
\centering\tabfont
\begin{tabular}{@{}llrrrr@{}}
\toprule
\textbf{Condition} & \textbf{Fails if, per serving} & \textbf{People} & \textbf{Option foods fail / pass / unknown} & \textbf{Fail share} & \textbf{Sole driver} \\
\midrule
Hypertension & sodium $\geq$460\,mg & 338 & 116 / 187 / 17 & 0.286 & 630 \\
Diabetes & carbohydrate $\geq$55\,g & 120 & 16 / 284 / 20 & 0.083 & 34 \\
High cholesterol & saturated fat $\geq$4\,g & 274 & 119 / 190 / 11 & 0.343 & 610 \\
Obesity & energy $\geq$400\,kcal & 246 & 28 / 260 / 32 & 0.115 & 168 \\
\midrule
\multicolumn{5}{@{}l}{Option slots failing two or more of the person's rules} & 182 \\
\multicolumn{5}{@{}l}{All failing option slots} & 1{,}624 \\
\bottomrule
\end{tabular}

\caption{Recommendation rule coverage on the test split. People: test persons with the condition.
Option foods: distinct option foods that fail, pass, or have an unknown outcome under the rule. Fail
share: share of option slots failing the rule among persons with the condition. Sole driver: failing
option slots that fail only this rule.}
\label{tab:app-rule-fail-dev}
\end{table}

\subsection{Inputs and Outputs}

The interfaces place different demands on grounding and evidence retrieval
(Table~\ref{tab:conditions}). VQA provides an image, question, answer space, and candidate records.
Recommendation provides option images and the person's profile, conditions, limits, and first-recall
intake but withholds option names and nutrient
records. Substitution provides a source image, possible source records, candidate replacements,
a goal, and limits.

We supply correct identities in the true-food condition to examine decisions with grounding
given, while retaining the original images. In recommendation, this condition also supplies the corresponding nutrient records. The names-plus-records condition removes image
content and supplies true identities with graph evidence instead. Recommendation's names-only
condition withholds both nutrient records and intake history, whereas names plus records includes
both. Its effects therefore concern joint access to nutrient and behavioral evidence; they do not
isolate nutrient records alone.

We request typed answers or complete rankings from direct models, together with image-to-record
assignments for VQA. Matching pipelines predict food identities, then apply the rules or use a text
reader to answer from the selected records. These structured outputs allow the evaluator to score decisions
independently of explanatory wording. Appendix~\ref{sec:app-prompts} reproduces the prompt templates.

\begin{table}[t]
\centering\tabfont
\begin{tabular}{@{}>{\raggedright\arraybackslash}p{0.14\linewidth}>{\raggedright\arraybackslash}p{0.19\linewidth}>{\raggedright\arraybackslash}p{0.18\linewidth}>{\raggedright\arraybackslash}p{0.18\linewidth}>{\raggedright\arraybackslash}p{0.18\linewidth}@{}}
\toprule
\textbf{Condition} & \textbf{Purpose of the control} & \textbf{\tvqa} & \textbf{\trec} & \textbf{\tsub} \\
\midrule
\rowcolor{groupgray}\multicolumn{5}{@{}l}{\textit{Standard input conditions}} \\
Full & n/a & image(s), question, records of the two named foods & eight option images, the person's conditions, rules and first-recall foods & image, goal and limits, records of the two possible foods and of eight candidates \\
Image removed & what the image contributes & gray canvas for each image & --- & --- \\
Records removed & what the graph input contributes & the two records withheld & the first-recall foods withheld & --- \\
Order permuted & sensitivity to listing order & the two images listed in the other order & --- & --- \\
\rowcolor{groupgray}\multicolumn{5}{@{}l}{\textit{Privileged conditions}} \\
True food given\priv & effect of supplying true-food evidence & true identity with image and records retained & identities and records of all eight options with images retained & true identity with image and records retained \\
Name only\priv & information available from food names & --- & true names and dietary constraints without images, nutrient records, or intake history & --- \\
Names + records\priv & effect of adding graph evidence & true names and records without image & true names, nutrient records, dietary constraints, and first-recall intake without images & true name and records without image \\
\bottomrule
\end{tabular}

\caption{Direct-model input conditions. \priv{} marks privileged information. n/a denotes an undefined
condition, and an em dash denotes a condition not evaluated for that task.}
\label{tab:conditions}
\end{table}

Input removal has different meanings across these interfaces. Removing records from a
retrieval-based system would also remove gallery and graph access, so we do not evaluate that
condition. Supplying true identities bypasses retrieval and reduces matching-plus-rules pipelines
to the true-food reference. For direct recommendation models, removing records instead withholds
first-recall history. These differences constrain comparisons across architectures: an input
control does not necessarily remove the same evidence from every system.

\subsection{Construction Checks}
\label{sec:app-checks}

We check that applying the rules reproduces every gold label and that paired instances have the
intended contrasting answers. These checks validate the implementation of the
record-based rules, not the visual association itself. Model inputs omit gold food bindings, targets, file paths, image source addresses, and
identifiers that could reveal labels. Training graph views exclude evaluation
respondents, and later-recall intake is available only to the scorer. File hashes, perceptual
hashes, and image-feature similarity check overlap across query, gallery, task, and split boundaries.

A model can produce the correct rule-based answer without necessarily using all intended evidence. We therefore examine whether public metadata, frequency, option order, or image text provides predictive shortcuts.
The probes
fit hidden labels from other clusters and are distinct from zero-shot task systems. After
examining metadata associations, we retained the probes as diagnostics rather than applying
the originally specified exclusion policy. No instance was excluded on this basis.

Image-source metadata predicts recommendation rule outcomes with AUC 0.523, relevance with 0.510,
and preference with 0.523. Served-image metadata probes reach AUC as high as 0.565 and VQA accuracy
up to 55.0\%; adding metadata to image-blind priors raises pooled VQA accuracy by at most
1.4 percentage points. These residual associations qualify claims about shortcut resistance;
the checks do not establish independence or rule out untested shortcuts.

\section{Evaluation and Implementation}
\label{sec:app-protocol}
\label{sec:app-stats}

\subsection{Scoring and Uncertainty}
\label{sec:app-design-scoring}

We first validate each output and then score its task-specific decision. For recommendation,
we distinguish nutrient-rule compliance from intake-derived relevance.

\paragraph{Output validation.} We validate each final prediction against the public instance schema after its system's parser
and normalizer. A
prediction is invalid if it duplicates or invents identifiers, omits an answer or ranking, returns an
answer outside the specified space, or links to an unknown image. We record the model's stated status
separately. Missing options in a nonempty ranking follow the tie policy below. Final invalid outputs count as incorrect and remain in the evaluation denominator. We retry an
infrastructure failure once before recording a runtime error, which is scored as invalid. Retaining these
failures in the denominator evaluates the complete system under its frozen inference settings. All scores
can be reproduced from stored predictions without further model inference.

\paragraph{\tvqa{} metrics.} We compute question accuracy by exact matching for yes/no and
choice answers, with a 5\% relative tolerance for nutrient amounts. A tie within 5\% is a valid choice answer. Non-answer statuses, including abstention,
score zero. Pair accuracy requires correct answers to both questions in a minimal pair, testing whether
the answer changes correctly when the image changes. Match accuracy separately measures the proportion of
images assigned their correct food identity in listed order. We accept food codes or record identifiers
in either the links field or a textual links block. Images without a stated assignment count as
unmatched. The historical text-only Match statistic in Table~\ref{tab:t1-full} instead conditions
on outputs containing links; for Qwen3-8B it covers 3{,}485 of 3{,}496 image cards. This denominator
difference concerns the diagnostic Match column, not question or pair accuracy.

The image-blind VQA prompts retain instance-specific opaque record identifiers. Paired questions
therefore share their task content and record order, but their literal prompts are not identical;
the small nonzero pair scores do not demonstrate visual grounding.

\paragraph{\trec{} metrics.} S1 is the proportion of
requests whose top two options both pass the person's rules (6/28 at random). S2 is the proportion
selecting both relevant options among requests whose top two both pass (1/6 at random). Request accuracy
is S1 times S2 (1/28 at random), placing joint success on a common denominator even when systems have different compliant subsets. Pair accuracy requires both requests of a person pair to be correct
(1/784 at random). Safety AUC is the probability that a passing option ranks above its same-category
failing alternative. Preference AUC is the probability that an option from the person's later-recall
categories ranks above an option from the paired person's later-recall categories, including both passing
and failing options. Both AUC measures give tied scores half credit. Safety AUC therefore measures rule
compliance in the ranking, not clinical safety.
Preference AUC measures discrimination of the benchmark's relevance labels, not an expressed food
preference.

The prompts request all eight options. We apply the same ranking rule to all generated outputs:
a nonempty list of distinct valid identifiers is scored as returned, with omitted options tied
below every listed option. We do not invent an order for the omitted options. Rankings containing
duplicate or unknown identifiers are invalid and score zero. Exact ties are scored by their
uniform expectation over possible orders. For seeded rows, S2 is
the ratio of Request totals to S1 totals pooled across the three stored seeds, consistent with its
reported uncertainty summaries. Other seeded centers are arithmetic means. Every main-table system is
evaluated on all 406 test requests. S2 nevertheless uses a system-specific denominator because it
includes only requests for which that system satisfies S1. Request accuracy retains the full denominator for comparison across systems.

\paragraph{Uncertainty.} Questions and requests can share visual inputs, so resampling them
individually would disregard that dependence. We instead resample image clusters: image
groups for \tvqa{}, option-image sets for \trec{}, and source-image pairs for \tsub{}.
Single-run SE is the standard deviation of estimates from 2{,}000 clustered bootstrap resamples.
Displayed intervals are 95\% pointwise percentile intervals from the same resamples; paired
effects use differences resampled within the same clusters. These intervals are descriptive; the paper
reports no formal hypothesis tests or multiple-comparison claims.

Trained models and the sampled direct Llama VQA/substitution controls report three-seed means
and SDs, except that seeded S2 pools Request/S1 totals and
its SD describes the three seed-specific ratios. Seeded effect intervals condition on the stored
seeds; they do not represent uncertainty over all possible training runs. The reported spread also does not cover model selection or collection bias.

\subsection{Models and Settings}
\label{sec:app-models}

We evaluate zero-shot language models, matching-and-reading pipelines, and trained intake
rankers (Table~\ref{tab:app-roster-dev}). Language models use supplied or retrieved
evidence, whereas rankers also learn behavioral associations from training respondents.

The trained baselines use factorization, graph propagation, or contrastive objectives to model
intake-derived relevance. They include matrix factorization with BPR~\cite{rendle2009bpr} and graph
encoders using GCN~\cite{kipf2017gcn}, GraphSAGE~\cite{hamilton2017graphsage}, and
GAT~\cite{velickovic2018gat} propagation. These encoders learn node-ID embeddings. GCN and
GraphSAGE average layer outputs; GAT uses its final-layer representation after learned
feature transformations and neighborhood attention.
Collaborative graph models include NGCF~\cite{wang2019ngcf} and LightGCN~\cite{he2020lightgcn},
while SGL~\cite{wu2021sgl}, SimGCL~\cite{yu2022simgcl}, and LightGCL~\cite{cai2023lightgcl}
use contrastive learning. Food-specific models include HFRS-DA~\cite{forouzandeh2024hfrsda},
MOPI-HFRS~\cite{zhang2025mopihfrs}, and RecipeRec~\cite{tian2022reciperec}.
LightGCN variants additionally use image, shuffled-image, or nutrition features. For LightGCL,
MOPI-HFRS, RecipeRec, and HFRS-DA, the main table uses the repository- and paper-based
implementations described below. Our task-matched adaptations of these models serve as supplementary controls. All use the
same benchmark respondents, food matches, and evaluator; training and person representations differ.

\paragraph{Decoding and budgets.} All runs use greedy decoding, except the text-only Llama-3.1-8B
VQA and substitution controls, which sample with temperature 0.6 and top-$p$ 0.9 over three seeds.
Non-reasoning vision--language models generate at most 256 new tokens and text-only baselines at
most 512. Reasoning-mode models use 4{,}096 tokens in \tvqa{} and in the recommendation input
controls (Table~\ref{tab:app-ladder}), and 8{,}192 tokens in the main recommendation rows and in
\tsub{}. The Qwen3-VL-8B matcher returns its assignments within 128 tokens. All budgets were
selected on the validation split and fixed before test evaluation.

\paragraph{Images.} We decode images to RGB, remove metadata, and downscale the longer side to at
most 448 pixels without enlarging smaller images. InternVL3.5-8B processes each image as a single
448-pixel tile. Feature-based matchers and recommenders use features computed from the original
images.

\paragraph{Matchers.} Every matcher passes its selected records to the same nutrient rules, so
score differences reflect the food identities it supplies. DINOv2-L returns the food of the
nearest gallery image by cosine similarity. SigLIP2 matches images to food names (zero-shot), to
gallery images, or to the equal-weight average of both similarities (fused). In \tvqa{}, matching
is restricted to the supplied candidate records; in recommendation, it searches all
option-eligible foods. Qwen3-VL-8B assigns VQA images to the candidate records and, in
recommendation, reranks each option's eight nearest DINOv2 gallery foods. SigLIP2 zero-shot
matches also supply the entry foods for Graph-RAG and for the substitution heads. In
recommendation, options predicted to pass rank before unknown and failing options, and a
naive-Bayes model fitted on training releases orders options within each group by predicting
later-recall categories from first-recall categories. The true-food reference uses the same
procedure with correct identities.

\paragraph{Graph-RAG readers.} Llama-3.1-8B and Qwen3-8B receive the same food identities and no
images: SigLIP2 zero-shot matches under full input, or true identities in the privileged
condition. Qwen3-8B runs with its reasoning mode off, except in rows marked (R), which use an
8{,}192-token reasoning budget (16{,}384 for KAPING in \trec{}). \emph{Records} verbalizes each
food's one-hop neighborhood as triples without truncating nutrient values. KAPING ranks triples
with all-MiniLM-L6-v2 embeddings~\citep{reimers2019sbert,wang2020minilm} and keeps the top 10 per
food in \tvqa{} and the top 6 per option, plus ten person triples, in \trec{}. Zero-shot CoT reasons
for up to 2{,}048 tokens before answer extraction; in \tsub{}, CoT (Qwen) uses 8{,}192 tokens
(Table~\ref{tab:t3-main}) and CoT (Llama) 2{,}048 (Table~\ref{tab:t3-full}). Think-on-Graph
searches from the matched foods to depth 3 with width 3. Recommendation readers also receive the
person's profile, nutrient rules, and first-recall history. All behavioral relations come from
training releases.

\paragraph{Trained recommenders.} The eight generic rankers in Table~\ref{tab:t2-main} (MF-BPR,
GCN, GraphSAGE, GAT, NGCF, LightGCN, SGL, and SimGCL) are trained on a graph of training
respondents and foods, with person--food edges weighted by log intake counts. Each person is
represented by the mean embedding of their first-recall foods (fold-in), and options are ranked
by inner product. The training objective scores a person's later-recall categories against those
of a paired training person, with negatives drawn from option-eligible foods of the same
popularity stratum. One hyperparameter grid (learning rate, L2 weight, embedding size, depth)
serves all configurations; we select each model's setting by preference AUC on a held-out 10\%
of training persons and train with early stopping (patience 5, at most 300 epochs). The encoders
follow their published propagation rules with minor simplifications for this objective.
Three LightGCN controls add image, shuffled-image, or nutrition features to the food nodes.

The food-specific baselines use repository implementations of LightGCL, MOPI-HFRS, and RecipeRec,
and a paper-based implementation of HFRS-DA, whose release lacks a trainable ranker. They share
the training respondents, SigLIP2 matches, rule filter, and scorer of the generic rankers but keep
their own objectives: LightGCL and RecipeRec use repository hyperparameters and final-epoch
checkpoints, and MOPI-HFRS combines ranking, health, and diversity losses with
MGDA~\citep{sener2018mgda}, selecting checkpoints by held-out-training recall plus health score.
For unseen respondents, LightGCL, RecipeRec, and HFRS-DA average first-recall food embeddings, and
MOPI-HFRS adds the respondent's features, foods, and condition-derived edges to its graph. We fix
implementation issues in the released MOPI-HFRS code and include all training respondents; all
changes were fixed before the reported runs. Table~\ref{tab:t2-adapted} reports versions of these
four models trained with the generic objective, and two health-aware variants of HFRS-DA and
MOPI-HFRS additionally use the person's conditions and first-recall foods at scoring (Table~\ref{tab:t2-full}). All recommenders use three seeds; we report
seed means and SDs, pooling S2 across seeds.

\paragraph{Compute.} Experiments run on a Slurm cluster with one NVIDIA A100 or A30 GPU per job.

\begin{table}[t]
\centering\tabfont
\begin{adjustbox}{max width=\linewidth}\begin{tabular}{@{}>{\raggedright\arraybackslash}p{0.25\linewidth}>{\raggedright\arraybackslash}p{0.40\linewidth}>{\raggedright\arraybackslash}p{0.29\linewidth}@{}}
\toprule
\textbf{Model} & \textbf{Checkpoint} & \textbf{Notes} \\
\midrule
\rowcolor{groupgray}\multicolumn{3}{@{}l}{\textit{Vision--language models (256 new tokens)}} \\
Qwen3-VL-2B & {\footnotesize\texttt{\detokenize{Qwen/Qwen3-VL-2B-Instruct}}} &  \\
Qwen3-VL-4B & {\footnotesize\texttt{\detokenize{Qwen/Qwen3-VL-4B-Instruct}}} &  \\
Qwen3-VL-8B & {\footnotesize\texttt{\detokenize{Qwen/Qwen3-VL-8B-Instruct}}} &  \\
InternVL3.5-8B & {\footnotesize\texttt{\detokenize{OpenGVLab/InternVL3_5-8B}}} &  \\
\rowcolor{groupgray}\multicolumn{3}{@{}l}{\textit{Vision--language models with a reasoning mode (task-specific token budgets)}} \\
GLM-4.1V-9B-Thinking & {\footnotesize\texttt{\detokenize{THUDM/GLM-4.1V-9B-Thinking}}} &  \\
MiMo-VL-7B-RL & {\footnotesize\texttt{\detokenize{XiaomiMiMo/MiMo-VL-7B-RL}}} &  \\
Qwen3-VL-8B-Thinking & {\footnotesize\texttt{\detokenize{Qwen/Qwen3-VL-8B-Thinking}}} & VQA and substitution only \\
\rowcolor{groupgray}\multicolumn{3}{@{}l}{\textit{Text-only language models (also the Graph-RAG readers)}} \\
Llama-3.1-8B & {\footnotesize\texttt{\detokenize{meta-llama/Llama-3.1-8B-Instruct}}} &  \\
Qwen3-8B & {\footnotesize\texttt{\detokenize{Qwen/Qwen3-8B}}} & reasoning mode off; on for (R) rows \\
\rowcolor{groupgray}\multicolumn{3}{@{}l}{\textit{Image matchers}} \\
DINOv2-L & {\footnotesize\texttt{\detokenize{facebook/dinov2-large}}} & food identity from the nearest gallery image \\
SigLIP2 so400m/14 & {\footnotesize\texttt{\detokenize{google/siglip2-so400m-patch14-384}}} & food identity from names, gallery images, or their fusion; not used in collection \\
\rowcolor{groupgray}\multicolumn{3}{@{}l}{\textit{Graph-embedding baselines (\tsub{} only)}} \\
TransE / RotatE / ComplEx / R-GCN & trained on foods excluded from evaluation instances & goal-conditioned decision head; SigLIP2 grounding. 3 seeds \\
\bottomrule
\end{tabular}
\end{adjustbox}
\caption{Evaluation scope and input settings.}
\label{tab:app-roster-dev}
\end{table}

\paragraph{Evaluation scope and model selection.}
\label{sec:app-ledger}
The image-selection checkpoints---Idefics3-8B, Qwen2.5-VL-7B, LLaVA-1.5-7B, and CLIP ViT-L/14---are
excluded from evaluation. Qwen3-VL is a separate release series and SigLIP2 replaces CLIP. These
exclusions do not eliminate collection bias or possible pretraining exposure.

The VQA matcher is evaluated in both image orders because its prompt contains both images.
Scoring accepts card-keyed record identifiers in VQA links and preserves explicit answer
delimiters, including boxed substitution answers. All generated recommendation rankings follow
the shared identifier and tie policy. Substitution outputs follow the completion rule in
Table~\ref{tab:t3-direct}. We retain predictions, configurations, and scoring records so these
procedures can be checked without repeating model inference.

All design and configuration choices were made without test data. Reporting S1 separately from
S2, image preprocessing, matcher choice, reasoning budgets, and task-matched training settings
were selected on the validation split or on held-out training respondents and foods, and were
frozen before test evaluation. The test split was used only for the final reported runs.
Diagnostic analyses focus on models chosen by their main-table scores; this choice affects
which analyses we report, not any model configuration.

\section{Input and Ranking Controls}
\label{sec:app-more-results}

Table~\ref{tab:t1-full} compares VQA performance as the supplied evidence changes.
Table~\ref{tab:t2-full} evaluates trained relevance ranking with food identities supplied.
Both diagnose components of the main results rather than rank complete systems again.

\begin{table}[t]\centering\tabfont\setlength{\tabcolsep}{3pt}
\begin{adjustbox}{max width=\linewidth}
\setlength{\tabcolsep}{1.5pt}
\begin{tabular}{@{}>{\raggedright\arraybackslash}p{150pt}ccccc@{}}
\toprule
\textbf{System} & \textbf{Image removed} & \textbf{Records removed} & \textbf{True food given\priv} & \textbf{Match} & \textbf{Invalid} \\
\midrule
\rowcolor{groupgray}\multicolumn{6}{@{}p{\linewidth}}{\textit{Match each image to a food, then apply the rules}} \\
DINOv2 match + rules$^{\circ}$ & n/a & n/a & $+$0.278\pmse{0.024} & 0.773\pmse{0.020} & 0.000 \\
SigLIP2 zero-shot match + rules$^{\circ}$ & n/a & n/a & $+$0.078\pmse{0.014} & 0.938\pmse{0.011} & 0.000 \\
SigLIP2 gallery match + rules$^{\circ}$ & n/a & n/a & $+$0.316\pmse{0.023} & 0.750\pmse{0.018} & 0.000 \\
SigLIP2 fused match + rules$^{\circ}$ & n/a & n/a & $+$0.173\pmse{0.019} & 0.862\pmse{0.015} & 0.000 \\
Qwen3-VL-8B match + rules & n/a & n/a & $+$0.143\pmse{0.016} & 0.877\pmse{0.014} & 0.000 \\
\rowcolor{groupgray}\multicolumn{6}{@{}p{\linewidth}}{\textit{Retrieve records for the SigLIP2-matched food, then a language model answers (Graph-RAG)}} \\
Retrieved records + Llama-3.1-8B & n/a & n/a & $+$0.059\pmse{0.011} & 0.938\pmse{0.011} & 0.000 \\
Retrieved records + Qwen3-8B & n/a & n/a & $+$0.057\pmse{0.010} & 0.938\pmse{0.011} & 0.000 \\
Retrieved records + Qwen3-8B (R) & n/a & n/a & $+$0.078\pmse{0.014} & 0.938\pmse{0.011} & 0.000 \\
KAPING + Llama-3.1-8B & n/a & n/a & $+$0.062\pmse{0.012} & 0.938\pmse{0.011} & 0.000 \\
KAPING + Qwen3-8B & n/a & n/a & $+$0.065\pmse{0.012} & 0.938\pmse{0.011} & 0.000 \\
KAPING + Qwen3-8B (R) & n/a & n/a & $+$0.078\pmse{0.014} & 0.938\pmse{0.011} & 0.000 \\
Zero-shot CoT + Llama-3.1-8B & n/a & n/a & $+$0.088\pmse{0.011} & 0.938\pmse{0.011} & 0.001 \\
Zero-shot CoT + Qwen3-8B & n/a & n/a & $+$0.075\pmse{0.014} & 0.938\pmse{0.011} & 0.000 \\
Think-on-Graph + Llama-3.1-8B & n/a & n/a & $+$0.068\pmse{0.013} & 0.938\pmse{0.011} & 0.000 \\
Think-on-Graph + Qwen3-8B & n/a & n/a & $+$0.075\pmse{0.013} & 0.938\pmse{0.011} & 0.000 \\
\rowcolor{groupgray}\multicolumn{6}{@{}p{\linewidth}}{\textit{Vision--language models answering directly}} \\
Qwen3-VL-2B & $-$0.210\pmse{0.010} & $-$0.444\pmse{0.012} & $+$0.145\pmse{0.011} & 0.465\pmse{0.010} & 0.003 \\
Qwen3-VL-4B & $-$0.376\pmse{0.014} & $-$0.561\pmse{0.014} & $+$0.203\pmse{0.012} & 0.697\pmse{0.012} & 0.001 \\
Qwen3-VL-8B & $-$0.283\pmse{0.013} & $-$0.591\pmse{0.014} & $+$0.170\pmse{0.011} & 0.765\pmse{0.013} & 0.000 \\
InternVL3.5-8B & $-$0.136\pmse{0.010} & $-$0.428\pmse{0.014} & $+$0.240\pmse{0.010} & 0.697\pmse{0.013} & 0.002 \\
\rowcolor{groupgray}\multicolumn{6}{@{}p{\linewidth}}{\textit{Vision--language models answering directly, with a reasoning mode}} \\
GLM-4.1V-9B (R) & $-$0.489\pmse{0.013} & $-$0.791\pmse{0.013} & $+$0.114\pmse{0.011} & 0.742\pmse{0.012} & 0.007 \\
MiMo-VL-7B-RL & $-$0.675\pmse{0.014} & $-$0.716\pmse{0.014} & $+$0.147\pmse{0.012} & 0.779\pmse{0.012} & 0.000 \\
Qwen3-VL-8B (R) & $-$0.774\pmse{0.013} & $-$0.859\pmse{0.013} & $+$0.107\pmse{0.012} & 0.884\pmse{0.012} & 0.051 \\
\rowcolor{groupgray}\multicolumn{6}{@{}p{\linewidth}}{\textit{Text-only language models (no image)}} \\
Llama-3.1-8B$^{\circ}$ & n/a & n/a & $+$0.294\pmse{0.002} & 0.422\pmse{0.002} & 0.000 \\
Qwen3-8B$^{\circ}$ & n/a & n/a & $+$0.284\pmse{0.009} & 0.455\pmse{0.004} & 0.000 \\
\bottomrule
\end{tabular}

\end{adjustbox}
\caption{VQA input and grounding controls. Changes are altered-input minus full-input accuracy in
listed order; $=$ means unchanged input and $^\circ$ order-invariant input. Match and Invalid use
listed order; text-only Match conditions on outputs containing links.
$\pm$: clustered-bootstrap SE, except three-seed SD for Llama. True food given: correct identities supplied.
Graph-RAG uses shared SigLIP2 zero-shot links.}
\label{tab:t1-full}
\end{table}

\subsection{Evidence Access and Rule Filtering}
\label{sec:app-input-diagnostics}

VQA accuracy changes after image removal range from $-$0.136 for InternVL3.5-8B to
$-$0.774 for Qwen3-VL-8B (R); changes after record removal range from $-$0.428 to $-$0.859
(Table~\ref{tab:t1-full}). Missing answers
contribute to these changes: image removal leaves 0.656 of MiMo-VL-7B-RL and 0.811 of
Qwen3-VL-8B (R) questions unanswered. The changes therefore combine evidence use and output
completion. In the main VQA comparison, SigLIP2 matching with rules exceeds Qwen3-VL-8B (R) by
$+$0.049 [0.023, 0.076] in question accuracy, a descriptive paired difference.

In the main recommendation evaluation, direct models receive option images without nutrient records.
Supplying records and intake history alongside true names raises Qwen3-VL-8B
S1 from 0.305 to 0.906 (Table~\ref{tab:app-ladder}). Image and name-only conditions omit first-recall
intake; names plus records restores it. Thus the contrast changes both nutrient evidence and
history and measures compliance rather than relevance. These controls use a 4{,}096-token
reasoning budget. Usability is the answered share under true-name input; MiMo-VL-7B-RL falls below
one half and its surviving answers are not representative of complete-system performance.
With images and history in the main task, Qwen3-VL-8B selects two failing foods less often than
4B but one failing food more often. Either outcome fails S1, so fewer complete misses need not
yield higher compliance.

\begin{table}[tp]\centering\tabfont\setlength{\tabcolsep}{3pt}
\begin{tabular}{@{}lccccc@{}}
\toprule
\textbf{Model} & \textbf{Images} & \textbf{Names\priv} & \textbf{Names + records\priv} & \textbf{Records step} & \textbf{Usable} \\
\midrule
Qwen3-VL-2B & 0.172 & 0.202 & 0.192 & $-$0.010 {\scriptsize[$-$0.053, 0.034]} & 1.000 \\
Qwen3-VL-4B & 0.192 & 0.204 & 0.800 & $+$0.596 {\scriptsize[0.513, 0.675]} & 1.000 \\
Qwen3-VL-8B & 0.148 & 0.305 & 0.906 & $+$0.601 {\scriptsize[0.507, 0.690]} & 1.000 \\
InternVL3.5-8B & 0.202 & 0.182 & 0.773 & $+$0.591 {\scriptsize[0.524, 0.658]} & 1.000 \\
GLM-4.1V-9B (R) & 0.224 & 0.335 & 0.739 & $+$0.404 {\scriptsize[0.315, 0.486]} & 0.869 \\
MiMo-VL-7B-RL$^{\S}$ & 0.266 & 0.196 & 0.751 & $+$0.556 {\scriptsize[0.492, 0.616]} & 0.433 \\
\bottomrule
\end{tabular}
\caption{Recommendation S1 across input conditions for evaluated direct models.
Images: option images with conditions and limits but without first-recall intake, so it differs from
full-input S1 in Table~\ref{tab:t2-main}; reasoning models use a 4{,}096-token budget here. Records step: names plus records minus names, with 95\% clustered intervals. Usable:
share of usable outputs given true names. $^{\S}$: usable on fewer than half of requests in one condition.}
\label{tab:app-ladder}
\end{table}

\Needspace{415pt}
\subsection{Ranking with Grounding Supplied}
\label{sec:app-retrained}

\begin{wraptable}{r}{0.53\textwidth}\centering\tabfont\setlength{\tabcolsep}{4pt}
\begin{adjustbox}{max width=\linewidth}
\begin{tabular}{@{}lc@{}}
\toprule
\textbf{Ranker} & \textbf{True-food preference AUC\priv} \\
\midrule
\rowcolor{groupgray}\multicolumn{2}{@{}l}{\textit{Primary implementations}} \\
LightGCL & 0.523\pmse{0.002} \\
HFRS-DA & 0.505\pmse{0.015} \\
MOPI-HFRS & 0.507\pmse{0.003} \\
RecipeRec & 0.544\pmse{0.005} \\
\rowcolor{groupgray}\multicolumn{2}{@{}l}{\textit{Task-matched training and intake fold-in}} \\
MF-BPR & 0.550\pmse{0.005} \\
LightGCN & 0.563\pmse{0.001} \\
LightGCN + image & 0.575\pmse{0.006} \\
LightGCN + shuffled image & 0.536\pmse{0.002} \\
LightGCN + nutrition & 0.569\pmse{0.011} \\
NGCF & 0.543\pmse{0.010} \\
GCN & 0.555\pmse{0.007} \\
GraphSAGE & 0.560\pmse{0.003} \\
GAT & 0.525\pmse{0.009} \\
SGL & 0.558\pmse{0.007} \\
SimGCL & 0.553\pmse{0.008} \\
LightGCL & 0.526\pmse{0.005} \\
HFRS-DA & 0.563\pmse{0.007} \\
MOPI-HFRS & 0.563\pmse{0.007} \\
RecipeRec & 0.582\pmse{0.009} \\
\rowcolor{groupgray}\multicolumn{2}{@{}l}{\textit{Health inputs at scoring}} \\
HFRS-DA & 0.545\pmse{0.005} \\
MOPI-HFRS & 0.533\pmse{0.005} \\
\bottomrule
\end{tabular}

\end{adjustbox}
\captionsetup{justification=raggedright,singlelinecheck=false}
\caption{Preference AUC of unfiltered ranker scores with true foods supplied, over all eight options. Groups distinguish
repository models, task-matched models and adaptations, and variants with health inputs at scoring.
$\pm$: SD over three training seeds.}
\label{tab:t2-full}
\end{wraptable}

We supply correct food identities in Table~\ref{tab:t2-full} to evaluate relevance ranking
without retrieval errors.
The task-matched models retain settings selected on held-out training respondents; image,
shuffled-image, nutrition, and health-input variants are separate controls rather than
substitutions chosen by test scores. True-food preference AUC evaluates all eight options.
The task-matched RecipeRec adaptation reaches 0.582, while the other task-matched base architectures
range from 0.525 to 0.563. The four primary implementations range from 0.505 to 0.544.
This diagnostic concerns the intake-derived category labels, not complete preferences.

The aligned-versus-shuffled LightGCN image contrast changes S1 by $-$0.020 [$-$0.060, 0.021] and
true-food preference AUC by $+$0.038 [0.014, 0.062]. Under this configuration, aligned features
improve the observed relevance diagnostic, while the compliance effect remains uncertain. The
health-input variants change the inference interface as described in the model settings; they
are not evidence that the benchmark measures clinical outcomes.

\finishwrap
\Needspace{165pt}
\begin{wraptable}{r}{0.51\textwidth}
\centering\tabfont\setlength{\tabcolsep}{4pt}
\begin{adjustbox}{max width=\linewidth}
\begin{tabular}{@{}lccc@{}}
\toprule
\textbf{Adaptation} & \textbf{S1} & \textbf{S2} & \textbf{Request} \\
\midrule
LightGCL & 0.597\pmse{0.015} & 0.199\pmse{0.014} & 0.119\pmse{0.011} \\
HFRS-DA & 0.569\pmse{0.036} & 0.224\pmse{0.004} & 0.128\pmse{0.010} \\
MOPI-HFRS & 0.543\pmse{0.008} & 0.213\pmse{0.018} & 0.115\pmse{0.008} \\
RecipeRec & 0.568\pmse{0.010} & 0.251\pmse{0.016} & 0.143\pmse{0.009} \\
\bottomrule
\end{tabular}
\end{adjustbox}
\captionsetup{justification=raggedright,singlelinecheck=false}
\caption{Task-matched adaptations of four recommenders in Table~\ref{tab:t2-main},
using the same SigLIP2 matches and rule filter. $\pm$: SD over three seeds; S2 pools seeds.}
\label{tab:t2-adapted}
\end{wraptable}

\paragraph{Task-matched adaptations.}
Table~\ref{tab:t2-adapted} supplements the main results with our adaptations trained
to predict later-recall categories. These use the common training procedure described in
Appendix~\ref{sec:app-models} and preserve each architecture's selected configuration.
Their architecture and training changes are evaluated together, so differences from the main
rows cannot be attributed to the objective alone. The RecipeRec adaptation reaches Request
accuracy 0.143, compared with 0.106 for its main implementation; MOPI-HFRS's adaptation instead
has lower S1, 0.543 versus 0.623. These contrasts further separate rule compliance from
intake-derived relevance.

\finishwrap
\subsection{Record Use and Output Completion}
\label{sec:app-analysis}

We evaluate record use on questions with correct SigLIP2 matches. Readers other than CoT answer amounts with accuracy of
at least 0.999; comparison accuracy ranges from 0.559 for Records (Qwen) to 0.996 for
KAPING (Qwen, R). Applying the rules answers every question in this subset correctly, so the readers'
remaining errors occur despite correct food matching. This subset excludes matching errors and
does not estimate full-task accuracy.

\subsection{Preference Prediction with True Food Identities} \label{sec:app-preference}

We evaluate behavioral references using correct food identities and the rule filter. This
removes grounding errors and tests whether observed intake history distinguishes relevant
categories among the passing options. Repeating
first-recall categories, with ties ordered by the training category prior, reaches request accuracy
of 0.234 [0.196, 0.275]. The diagnostic naive Bayes
reference tests a learned category transition using half of the adult training respondents with
reliable paired recalls; it reaches
0.212 [0.177, 0.249], against chance of 0.167 after true-food rule filtering. To retain the pipeline comparison, the true foods + rules reference instead uses its unchanged
naive Bayes estimator, fitted to all training respondents with reliable paired recalls, and
orders the true-food passing options from the person's intake history. Its request accuracy is
0.192 [0.160, 0.224]. Supplying true foods removes grounding errors from these
comparisons, but does not make any of the specified estimators an upper bound on personalization.

For these references, preference AUC assesses the ordering of all four rule-passing options. The behavioral references obtain AUCs of 0.576 and 0.568,
respectively. These estimates describe the selected history-based estimators, not the information recoverable
by every possible ranking method. The learned-ranker AUC in Table~\ref{tab:t2-full} uses all eight options and is not directly
comparable to the four-option behavioral-reference AUCs.

\subsection{Generation Limits in Input Controls} \label{sec:app-control-output}

Input controls can affect response completion as well as correctness. We report output validity
alongside accuracy to distinguish these effects. On single-food VQA with true
identities, Qwen3-VL-8B (R) scores below the other leading models because 0.024 of its outputs reach the
4{,}096-token limit. Accuracy on the remaining questions is 0.997. The lower aggregate score therefore
includes output failures, not only errors in reading nutrient records.

Recommendation controls show a similar distinction. MiMo-VL-7B-RL fails the usability check because 0.562
of its true-name outputs reach the 4{,}096-token budget, leaving 0.433 usable. With names plus records,
0.254 of GLM-4.1V-9B (R) outputs are truncated. Nevertheless, it satisfies S1 on 300 of its 301 answered
requests. These results distinguish complete-system performance from correctness on a model-specific
subset of usable responses. The latter describes surviving outputs, rather than an alternative score on
the full benchmark.

\section{\tsub{} Details}
\label{sec:app-t3}

\paragraph{Task inputs.} Each instance pairs a source-food image with a condition-specific
reduction goal: sodium for hypertension, saturated fat for high
cholesterol, carbohydrate for diabetes, and energy for obesity. Each of the person's conditions, including the goal condition, imposes a
per-serving limit: sodium below
460\,mg, saturated fat below 4\,g, carbohydrate below 55\,g, and energy below 400\,kcal.
The graph provides records for two possible source foods and eight replacements
from the same food family. A food family is a WWEIA category or a predefined group of related
categories, such as regular and diet soft drinks. The released family registry lists the food
codes in each group. The two source images are dissimilar (DINOv2 cosine below 0.6) and show foods
with different code prefixes and names. Each record gives the name, food code, serving weight, and at most eight
ingredients. It also lists per-serving energy, carbohydrate, sugars, saturated fat, and sodium.

\paragraph{Decision rule.} We define a valid replacement as a candidate that contains at least 20\% less of the goal nutrient per serving
than the source food, increases none of the other constrained nutrients (energy, carbohydrate,
saturated fat, sodium) by more than 25\%, and satisfies every personal limit. The system selects a
qualifying candidate or returns ``none'' if no candidate meets all conditions. The decision thus
combines relative improvement over the source with absolute nutrient-rule compliance.

\paragraph{Gold labels.} Applying the rule to FNDDS 2017--2018 serving-level values,
we require each candidate to have an unambiguous outcome for both possible source
foods, with margins around every threshold. We also recompute the source-relative clauses per
100\,g, retaining the same per-serving personal limits, and require candidate eligibility to agree.
An instance contains
either a unique valid replacement or none. Rule execution with the true source identity therefore
scores 1.000 by construction. This consistency check establishes the intended rule outcome;
it does not validate culinary equivalence or clinical suitability. Human annotators have not assessed replacement
plausibility.

Each substitution pair shares the candidate pool and text but supplies two source images that
require different answers, never both \emph{none}. The test split contains 610 pairs. Construction balances goals and answer roles, limits concentration
within food families, and excludes image overlap with other tasks and the gallery. Some
distractors meet the nutrient-reduction goal but violate another constraint, so the lowest
target-nutrient value alone need not identify a valid replacement.

Pair accuracy is the primary metric because a deterministic image-blind system receives identical
inputs for both members and cannot answer both correctly. Instance accuracy, true-identity
accuracy, and invalid-output rate provide complementary diagnostics. To assess whether learned food representations add useful information beyond nutrient values,
the graph-embedding baselines use TransE, RotatE, ComplEx, or
R-GCN~\cite{bordes2013transe,sun2019rotate,trouillon2016complex,schlichtkrull2018rgcn} with a
goal-conditioned head trained on rule-labeled food pairs outside the evaluation food sets.
The systems in Table~\ref{tab:t3-main} L2-normalize each graph vector before fusing it with
the food's five per-serving nutrient values~\citep{kristiadi2019literale} and passing it to the head.
We pretrain TransE, ComplEx, and RotatE on the static food graph, which includes evaluation
foods, and fit the fusion and decision head on rule-labeled pairs excluding evaluation foods.
The encoder remains fixed during head fitting; R-GCN instead learns its encoder jointly with
the head. We choose configurations and checkpoints using held-out training-food loss, then
freeze them before test evaluation. All systems use SigLIP2 zero-shot source matches and the
same personal-limit filter.

\paragraph{Encoder and head training.}
\label{sec:app-t3-training-audit}
We use the authors' RotatE implementation and source-equivalent PyTorch ports of the original
TransE and ComplEx releases. The ports preserve initialization, losses, corruption procedures,
and parameter updates; they are checked against the released operations. TransE uses separate
head and tail corruptions, margin loss, SGD, and unit-norm entity projection. ComplEx uses the
released corruption sampler, logistic loss, L2 regularization, and AdaGrad. RotatE uses filtered
head/tail corruption and self-adversarial negative sampling with Adam.

The selected TransE setting has dimension 50, L2 distance, margin 0.5, and learning rate 0.01;
ComplEx has complex dimension 150, regularization 0.03, learning rate 0.5, and one negative per
positive; RotatE has complex dimension 500, margin parameter 6, adversarial temperature 0.5,
and learning rate $5\times10^{-5}$. We compare two source-backed settings per encoder using
held-out training-food loss. Encoder schedules run up to 500 epochs for TransE, 1{,}000 for
ComplEx, and 80{,}000 updates for the selected RotatE setting, retaining the best complete
encoder--head checkpoint. At each checkpoint assessment, we fit the common head for at most
40 epochs with patience 5 on held-out binary cross-entropy. A fixed probability threshold of
0.5 determines candidate eligibility; no threshold is selected on test results.

The nutrient-values-only control retains the head and nutrient inputs but removes the graph
vector. The comparison assesses the added representation within this supervised interface. We filter
outputs by the person's limits and report the mean and standard deviation over 3 training seeds, rather than selecting the best seed. Every graph embedding scores below the control in pair accuracy
(Table~\ref{tab:t3-main}), and below deterministic rules using the same SigLIP2 matches.
Thus graph pretraining does not improve complete-rule decisions in these evaluated systems;
this comparison does not isolate encoder architecture from its optimization and fitted head.

Tables~\ref{tab:t3-full} and~\ref{tab:t3-direct} compare readers with the correct source identity
supplied and report output validity and truncation. These controls distinguish errors that persist
after source identification from failures to return a usable decision. Supplying the true source bypasses image matching while retaining the
nutrient records. This control measures record use when source identification is already correct;
the main task requires identifying the source from its image.
Table~\ref{tab:t3-full} omits CoT (Qwen); Table~\ref{tab:t3-main} reports its substitution
configuration, whose 8{,}192-token budget was fixed before testing.

\begin{table}[t]
\centering\tabfont
\setlength{\tabcolsep}{3pt}
\begin{tabular}{@{}lcccc@{}}
\toprule
\textbf{Reader} & \textbf{Pair} & \textbf{True-source pair\priv} & \textbf{Invalid} & \textbf{Truncated} \\
\midrule
Records (Llama) & 0.013\pmse{0.005} & 0.013\pmse{0.005} & 0.000 & 0.000 \\
Records (Qwen) & 0.005\pmse{0.003} & 0.011\pmse{0.004} & 0.000 & 0.000 \\
Records (Qwen, R) & 0.861\pmse{0.014} & 0.916\pmse{0.011} & 0.016 & 0.016 \\
KAPING (Llama) & 0.020\pmse{0.006} & 0.015\pmse{0.005} & 0.000 & 0.000 \\
KAPING (Qwen) & 0.008\pmse{0.004} & 0.015\pmse{0.005} & 0.000 & 0.000 \\
KAPING (Qwen, R) & 0.793\pmse{0.016} & 0.867\pmse{0.014} & 0.034 & 0.034 \\
CoT (Llama) & 0.110\pmse{0.013} & 0.085\pmse{0.011} & 0.002 & 0.041 \\
ToG (Llama) & 0.011\pmse{0.004} & 0.015\pmse{0.005} & 0.001 & 0.000 \\
ToG (Qwen) & 0.005\pmse{0.003} & 0.008\pmse{0.004} & 0.000 & 0.000 \\
\bottomrule
\end{tabular}

\caption{Substitution reader controls with SigLIP2 matches. True-source pair: correct source identity
supplied. $\pm$: clustered-bootstrap SE. Invalid and Truncated are shares of all instances.
CoT (Llama) uses 2{,}048 reasoning tokens.}
\label{tab:t3-full}
\end{table}

\begin{table}[t]
\centering\tabfont
\setlength{\tabcolsep}{3pt}
\begin{tabular}{@{}lcccc@{}}
\toprule
\textbf{System} & \textbf{Pair acc.} & \textbf{True-food pair\priv} & \textbf{Invalid} & \textbf{Truncated} \\
\midrule
\rowcolor{groupgray}\multicolumn{5}{@{}l}{\textit{Direct vision--language models}} \\
Qwen3-VL-2B & 0.010\pmse{0.004} & 0.008\pmse{0.004} & 0.115 & 0.000 \\
Qwen3-VL-4B & 0.031\pmse{0.007} & 0.031\pmse{0.007} & 0.006 & 0.000 \\
Qwen3-VL-8B & 0.023\pmse{0.006} & 0.038\pmse{0.008} & 0.002 & 0.000 \\
InternVL3.5-8B & 0.013\pmse{0.005} & 0.025\pmse{0.006} & 0.189 & 0.149 \\
GLM-4.1V-9B (R) & \underline{0.751}\pmse{0.018} & 0.944\pmse{0.009} & 0.010 & 0.010 \\
MiMo-VL-7B-RL & \textbf{0.779}\pmse{0.017} & 0.962\pmse{0.008} & 0.001 & 0.001 \\
Qwen3-VL-8B (R) & 0.644\pmse{0.020} & 0.820\pmse{0.016} & 0.134 & 0.134 \\
\rowcolor{groupgray}\multicolumn{5}{@{}l}{\textit{Text-only models with true food identity and records\priv}} \\
Llama-3.1-8B & 0.011\pmse{0.003} & 0.011\pmse{0.003} & 0.000 & 0.000 \\
Qwen3-8B & 0.020\pmse{0.006} & 0.020\pmse{0.006} & 0.000 & 0.000 \\
Qwen3-8B (R) & 0.934\pmse{0.010} & 0.934\pmse{0.010} & 0.007 & 0.007 \\
\bottomrule
\end{tabular}

\caption{Substitution evaluation of direct vision--language and text-only models. Text-only models
receive the true source identity and records. Invalid: no single parseable candidate identifier.
Unfinished free-text mentions do not count as final answers; a completed structured answer remains
scorable. Other columns follow Table~\ref{tab:t3-full}; $\pm$ is clustered-bootstrap SE, except
three-seed SD for Llama.}
\label{tab:t3-direct}
\end{table}

Qwen3-VL-8B (R) has 163 invalid substitution outputs, all truncated. Its instance accuracy is
0.918 on completed outputs and 0.795 overall. Accuracy on completed outputs excludes unfinished
cases and cannot predict performance with a larger budget. InternVL's low accuracy persists on
completed outputs, so truncation alone cannot explain its result.

We require a completed answer rather than a candidate mentioned in unfinished free text.
The same completion policy applies to every substitution system and retains explicitly parsed
structured answers. Boxed answer delimiters are preserved by the parser. No test run records
a runtime error. Qwen3-VL-2B exhibits a different failure:
all 140 invalid full-input outputs name a possible source record rather than a replacement.
Those answers reflect confusion between the two roles, rather than loss of a valid answer during parsing.

The paired-image design makes the source consequential, but does not establish that every other
input is necessary. We therefore also check how personal limits, the stated goal, and the candidate
pool constrain the answer. Replacing all condition limits with the goal condition's limit alone changes the gold
decision in only 0.022 of test instances. Sufficiently precise recalled nutrient values could also
support an answer without the supplied records. Moreover, candidate pools carry information about
the intended goal: the goal nutrient is recoverable from the pool and two possible source
records, without the stated goal, for 0.807 of instances. Metadata probes for the \emph{none} answer reach an AUC of
0.537 on pairs with one \emph{none} answer and 0.516 over all instances.
Answer roles are balanced globally rather than within each goal, and the sibling-prior check uses a mean. Together, these properties bound claims about
input dependence: the design makes source grounding consequential, but does not fully isolate the
effects of goals, records, and personal limits. Construction consistency also does not establish
clinical validity.

\section{Supporting Diagnostics}
\label{sec:app-why}

We use stored food links, rule outcomes, and answers to identify failures underlying the
aggregate scores. These observations do not establish internal reasoning or isolate every causal
contribution.

\subsection{VQA Grounding and Comparison Reading}
\label{sec:app-why-t1}

The matching-and-rules pipelines answer every correctly matched question, and their question
accuracy equals the share with all matches correct (Table~\ref{tab:app-error-types}). Matching both comparison images
to one record forces an incorrect tie: DINOv2 does this on 0.350 of comparisons and SigLIP2 on 0.096.
Given correct SigLIP2 links, native-thinking readers answer 0.999 of questions correctly. Records
(Qwen) instead answers comparisons with accuracy 0.559 and selects B in 0.642 of them.
KAPING and Think-on-Graph retrieve the requested nutrient for every matched food; missing that
nutrient therefore does not explain their comparison errors.

\subsection{Recommendation Rule Outcomes and Ranking}
\label{sec:app-why-t2}

All matchers use the same nutrient rules and naive-Bayes ranking. A
retrieved record can pass a rule even when the true food fails it: SigLIP2 maps
0.292 of truly failing options to passing records, versus 0.486 for DINOv2. Task-matched rankers and adaptations share
these predicted outcomes and change only within-group order. Their ability to distinguish
true from false passes is a different property from the intake-derived relevance target used
for training; the controlled comparison below separates these effects.

A margin diagnostic ranks the same stored matches by compliance rather than intake-derived
relevance. It orders them by the negative maximum nutrient-to-limit
ratio across stated rules, placing records further below those limits first. It reaches S1 0.714,
$+$0.155 [0.040, 0.270] above random within-group ordering. It is a stored-prediction diagnostic,
not an evaluated benchmark system or validated health measure. For relevance, complementary
paired-person labels make any common image-aligned order's preference AUC exactly 0.5.
History-based references are defined in Appendix~\ref{sec:app-preference}.

For MOPI-HFRS, rankings among predicted passing options correlate with training-food popularity
(mean Spearman $\rho=0.798$). Popularity counts distinct training respondents with reliable
intake of the matched food. We compute correlations within each request and seed, then average
over seeds and requests. The 366 evaluable requests have at least three predicted passes and
nonconstant rankings and popularity; undefined correlations are excluded. This association
describes the output ordering, without isolating why training produced it.
Its true-food preference AUC of 0.507 shows that this ordering provides little separation of
the paired respondents' intake-derived labels, even when food identification is supplied.

\paragraph{Retrieving and using nutrient evidence.}
We inspect ToG's returned evidence to ask whether its low recommendation accuracy is explained
by missing nutrient facts. Qwen retrieves 93.9\% of the facts required by the stated rules,
compared with 14.7\% for Llama. Coverage concerns the matched records and does not remove
food-identification errors. We count a fact only when it has a finite value in the required
unit, or a per-100-g value with a usable serving weight. All eight options have the required
nutrients in 343 of 406 Qwen requests but only 39 Llama requests. Yet complete-request accuracy
remains low for both (Table~\ref{tab:t2-main}). Missing nutrient retrieval alone therefore cannot
explain Qwen's result: having these values available does not ensure correct rule execution and
history-based ordering. Complete coverage of every option is not necessary to identify a valid
top two, so these counts describe evidence availability rather than a causal error attribution.

\Needspace{240pt}
\begin{wraptable}{r}{0.51\textwidth}
\centering\small
\begin{adjustbox}{max width=\linewidth}
\begin{tabular}{@{}lcc@{}}
\toprule
System & Relevant pair & Same choice \\
\midrule
GCN & $0.190\pm0.014$ & $0.269\pm0.038$ \\
LightGCN & $0.191\pm0.004$ & $0.228\pm0.006$ \\
SGL & $0.206\pm0.016$ & $0.217\pm0.015$ \\
LightGCL & $0.178\pm0.019$ & $0.246\pm0.060$ \\
HFRS-DA & $0.171\pm0.036$ & $0.358\pm0.032$ \\
MOPI-HFRS & $0.172\pm0.015$ & $0.675\pm0.068$ \\
\bottomrule
\end{tabular}
\end{adjustbox}
\captionsetup{justification=raggedright,singlelinecheck=false}
\caption{True-food, perfect-filter diagnostic. Relevant pair: both relevant foods selected per
request (406 requests). Same choice: identical top-two food sets across paired people (203 pairs),
whose correct sets are complementary. Means and sample SD over three seeds; ties are integrated
exactly, independently across paired requests. Lower same-choice rates are necessary for
separating these people but do not alone establish correct personalization.}
\label{tab:ranker-person-control}
\end{wraptable}

\paragraph{Separating filtering from ranking.}
We replay all twelve primary rankers' three-seed predictions and verify identical food matches
and predicted rule outcomes. Among predicted passes, MOPI-HFRS and HFRS-DA distinguish true
from false passes with AUC $0.567\pm0.012$ and $0.546\pm0.024$, respectively, compared with
$0.537\pm0.024$ for LightGCN and $0.489\pm0.004$ for GCN. This diagnostic uses the 298 requests
containing both kinds of predicted pass. It concerns compliance recovery, not relevance.

To test relevance separately, we restrict each ranker's stored true-food scores to the four
truly compliant options. The corrected rankers' true-food runs already apply this exact filter;
for the other rankers we apply it during rescoring. Neither operation changes their within-pass
order or retrains the model. We then compare paired requests with identical foods and rule
outcomes but complementary relevance labels. Table~\ref{tab:ranker-person-control} reports both
recovery of the two relevant foods and the frequency of identical selections across people.

MOPI-HFRS frequently gives the same answer where the task requires a change with the person.
Together with its popularity association and near-chance raw preference AUC, this locates its
weakness in person-specific ordering, rather than food compliance alone. HFRS-DA also repeats
selections more often than LightGCN, although its per-seed relevant-pair scores vary substantially.
These controls establish the observed decision pattern; the small accuracy gaps do not establish
statistical superiority or identify which training component produced it. Plain GCN has lower
main-table Request accuracy than MOPI-HFRS and HFRS-DA; its higher conditional S2 uses a different
subset of requests and should not be interpreted as overall superiority.

\subsection{Substitution Complete-Rule Decisions}
\label{sec:app-why-t3}

The complete substitution rule requires rejecting every candidate when none qualifies. This
occurs on 440 instances. A policy that never returns \emph{none} can solve only
0.279 of pairs even with the true source. CoT (Llama) returns \emph{none} correctly on 86 of these
instances; of its 1{,}032 candidate choices, 909 meet the goal clause but only 419 meet the
other-nutrient clause. Its low pair accuracy therefore accompanies observed complete-rule errors,
with little invalid output or truncation (Table~\ref{tab:t3-full}). Native-thinking Qwen3-8B configurations score $+$0.856 [0.828, 0.884] higher with retrieved
records and $+$0.915 [0.890, 0.938] higher as text-only readers. Thinking mode, token budget,
and answer prefill differ, so these contrasts do not isolate a single inference component. These effects describe the evaluated configurations, not model-family
superiority under matched inference settings.

To separate source identification from learned rule decisions, we compare each graph head's
image-linked predictions with the same checkpoint given the true source. Correct identities
raise instance accuracy from 0.953 to 0.984 for TransE, 0.942 to 0.973 for ComplEx, and
0.920 to 0.949 for RotatE. Similar gains of about three percentage points show that their
shared matcher accounts for part of the error, while the remaining gaps arise after grounding.

Most residual errors reject a valid substitute. Across 1{,}220 instances and three seeds,
TransE misses a valid replacement in 46 of its 59 remaining errors; the corresponding counts
are 83 of 100 for ComplEx and 127 of 188 for RotatE. These are pooled seed--instance decisions,
not independent examples. In each missed case, the gold candidate's learned score falls below
the fixed eligibility threshold. Supplying the right food record therefore does not ensure that
a learned head accepts every valid replacement. This result motivates studying rejection and
constraint satisfaction separately; it does not attribute the differences solely to graph
architecture, and the threshold is unchanged throughout the analysis.

\section{Access, Licensing, and Maintenance}
\label{sec:app-datasheet}
\label{sec:app-credits}

\bench{} is intended for benchmark research: nutrient values describe database food codes rather
than laboratory measurements of depicted meals, and the independently collected images do not
document respondents' actual consumption. These properties preclude treating the resource as
validated dietary or clinical decision support.

\paragraph{Dataset and benchmark dashboard.} The dashboard supports inspection of the aligned
resource before a researcher selects a retrieval or reasoning experiment. Its dataset card
documents coverage and access, and its searchable catalog exposes food categories, nutrient
profiles, and missing values. These views help identify which records contain the evidence
required by a proposed task. Result tables retain system inputs, metric denominators, and
uncertainty types so that researchers can distinguish changes in available evidence from
changes in scores. We release the dataset, task instances, evaluation code, and dashboard at
\url{https://umami-benchmark.github.io/}. Image files are distributed according to their
source licenses, as described below.

\paragraph{Research reuse.} The aligned records and paired tasks are intended for visual entity
linking, grounded language-model evaluation, and constrained ranking research. Testing a new
image encoder uses the image files or source links and the food-code links; comparing record
readers uses the supplied evidence and scoring rules; evaluating intake-based ranking uses the
released first-recall histories and candidate sets.

\paragraph{Distribution and license.} From the 20{,}492-image corpus, 2{,}040 images are
designated for file distribution under CC BY, CC BY-SA, or public-domain terms. This subset
includes 1{,}160 of the 8{,}821 gallery images (13.2\%). For other images with recoverable
sources, we provide source addresses, creators, and content hashes so that users can retrieve
and verify the original files. The 1{,}640 images without recorded source addresses include
830 gallery images. None belongs to the designated file-distribution subset; we release only
their hashes and food-code links. Hashes verify available files but cannot recover missing
ones, so these artifacts alone do not reconstruct the complete gallery. Each image retains its
original license. Annotations, questions, splits, and derived tables use CC BY 4.0, and the
code is released under the [MIT] license.

\paragraph{Image recovery.} Five gallery images were lost through file overwriting; two additional
gallery images and one recommendation validation image have perceptually similar restorations
rather than byte-identical originals. All current test query files match their recorded hashes.
The saved gallery embeddings were computed from the original files and remain available. Current
matchers use these embeddings, without loading gallery pixels into model prompts. Reproducing
the embeddings from images therefore requires recovery of the original files or a separately
versioned gallery and rerun; the saved-feature results do not establish reproduction from the
restored pixels.

\paragraph{Maintenance.} Versioned releases and checksum manifests identify the artifacts
used in published results. Corrections receive a new version, while earlier versions remain
available. Following a takedown request, we replace the hosted image with a source pointer.
Pointers may also become unavailable through link loss; where the source remains accessible,
content hashes allow users to detect changes to its files.

The unmodified dataset image in Figure~\ref{fig:food-record-example}, \emph{Spaghetti Bolognese with peas
and broccoli}, is by \href{https://commons.wikimedia.org/wiki/File:Liat_Portal_for_Foodie_Disorder_-_Spaghetti_Bolognese_with_peas_and_broccoli.jpg}{HaJunkiyada}
(Wikimedia Commons), licensed under \href{https://creativecommons.org/licenses/by-sa/4.0/}{CC BY-SA}.
We do not reproduce NoDerivatives images in modified layouts.

\section{Comparison with Related Resources}
\label{sec:app-related}
\providecommand{\cmpY}{\ensuremath{\checkmark}}\providecommand{\cmpP}{(\ensuremath{\checkmark})}
\providecommand{\cmpN}{\ensuremath{\times}}\providecommand{\cmpNA}{n/a}\providecommand{\cmpU}{?}
\begin{table}[t]\centering\tabfont\setlength{\tabcolsep}{2pt}\renewcommand{\arraystretch}{0.95}
\begin{adjustbox}{max width=\linewidth}
\begin{tabular}{@{}>{\raggedright\arraybackslash}p{100pt}>{\raggedright\arraybackslash}p{44pt}ccccccc>{\raggedright\arraybackslash}p{65pt}@{}}
\toprule
\textbf{Resource} & \textbf{Venue} & \shortstack{\textbf{Image}\\\textbf{input}} & \shortstack{\textbf{Image}\\\textbf{$\to$KB}} & \shortstack{\textbf{Nutr.}\\\textbf{KB}} & \textbf{Person} & \shortstack{\textbf{Held-out}\\\textbf{people}} & \textbf{Controls} & \shortstack{\textbf{Human}\\\textbf{val.}} & \textbf{Tasks} \\
\midrule
\textbf{\bench{} (ours)} & n/a & \cmpY & \cmpY & \cmpY & \cmpY & \cmpY & \cmpY & \cmpP & VQA, rec., substitution \\
\midrule
NGQA \cite{zhang2024nutrition} & ACL'25 & \cmpN & \cmpN & \cmpY & \cmpY & \cmpN & \cmpN & \cmpP & Nutrition QA \\
GLEN-Bench \cite{glenbench2026} & TMLR'26 & \cmpN & \cmpN & \cmpY & \cmpY & \cmpP & \cmpP & \cmpN & Risk, rec., QA \\
MOPI-HFRS \cite{zhang2025mopihfrs} & KDD'25 & \cmpN & \cmpN & \cmpY & \cmpY & \cmpN & \cmpN & \cmpN & Health-aware rec. \\
NutriBench \cite{hua2025nutribench} & ICLR'25 & \cmpN & \cmpN & \cmpY & \cmpN & \cmpN & \cmpP & \cmpP & Carb estimation \\
DiningBench \cite{jin2026diningbench} & ACL'26 & \cmpY & \cmpN & \cmpP & \cmpN & \cmpNA & \cmpP & \cmpY & Dish VQA, nutrition est. \\
FoodieQA \cite{li2024foodieqa} & EMNLP'24 & \cmpY & \cmpN & \cmpN & \cmpN & \cmpNA & \cmpP & \cmpY & Food VQA \\
WorldCuisines \cite{winata2025worldcuisines} & NAACL'25 & \cmpY & \cmpP & \cmpN & \cmpN & \cmpNA & \cmpP & \cmpP & Dish naming, origin \\
DietAI24 \cite{yan2025dietai24} & Commun.\ Med.'25 & \cmpY & \cmpY & \cmpY & \cmpN & \cmpNA & \cmpP & \cmpP & Code recognition, nutrient est. \\
Health MFQA \cite{hu2025healthmfqa} & TOMM'25 & \cmpY & \cmpP & \cmpP & \cmpP & \cmpNA & \cmpP & \cmpN & Health-suitability food VQA \\
Nutrition5k \cite{thames2021nutrition5k} & CVPR'21 & \cmpY & \cmpN & \cmpY & \cmpN & \cmpNA & \cmpP & \cmpN & Nutrient regression \\
OVEN \cite{hu2023oven} & ICCV'23 & \cmpY & \cmpY & \cmpN & \cmpN & \cmpNA & \cmpN & \cmpY & Entity recognition \\
InfoSeek \cite{chen2023infoseek} & EMNLP'23 & \cmpY & \cmpY & \cmpN & \cmpN & \cmpNA & \cmpY & \cmpY & Knowledge VQA \\
Encyclopedic-VQA \cite{mensink2023encyclopedic} & ICCV'23 & \cmpY & \cmpY & \cmpN & \cmpN & \cmpNA & \cmpY & \cmpY & Knowledge VQA \\
EHRXQA \cite{bae2023ehrxqa} & NeurIPS'23 & \cmpY & \cmpN & \cmpN & \cmpY & \cmpY & \cmpP & \cmpP & EHR QA (image, table) \\
pFoodReQ \cite{chen2021pfoodreq} & WSDM'21 & \cmpN & \cmpN & \cmpY & \cmpP & \cmpN & \cmpP & \cmpN & Constrained rec. QA \\
KERL \cite{mohbat2025kerl} & ACL'25 & \cmpN & \cmpN & \cmpY & \cmpP & \cmpNA & \cmpP & \cmpN & Rec., recipe gen. \\
HUMMUS \cite{bolz2023hummus} & RecSys'23 & \cmpN & \cmpN & \cmpY & \cmpP & \cmpN & \cmpN & \cmpN & Health-aware rec. \\
\bottomrule
\end{tabular}
\end{adjustbox}
\caption{Comparison with published resources. \cmpY{}: yes; \cmpP{}: partial; \cmpN{}: no; n/a: not
applicable. Image input and image-to-KB association are domain-independent; partial association denotes implicit or unscored linking.
Human validation for \bench{} covers image labels in the initial corpus, not the test-only extension or rule-generated answers.}
\label{tab:app-comparison}
\end{table}

Table~\ref{tab:app-comparison} compares how related resources link images, food records, and
person-specific decisions, together with their validation and controls. Held-out
people are excluded from training. Controls denotes evidence-input comparisons; partial indicates
restricted or confounded comparisons. Human validation for \bench{} concerns initial-corpus image
labels, excluding the extension and rule-generated task answers.

\section{Grounding Errors and Example Scope}
\label{sec:app-cases}

We group test errors by the stated grounding outcome in Table~\ref{tab:app-error-types} to
distinguish incorrect food assignments from subsequent record-use errors. The illustrations in
Appendix~\ref{sec:app-dataset-examples} instead explain the scope of the dataset's labels. Every incorrect answer from the DINOv2 and SigLIP2 rule pipelines follows an
incorrect match. On the SigLIP2 matches, Graph-RAG readers without reasoning also answer
0.040--0.128 of questions incorrectly despite correct matches, identifying observed record-use
errors separately from grounding failures. Direct-model errors occur with wrong, correct, or
unstated assignments; an unstated assignment does not establish what the model recognized internally.

The pasta example in Figure~\ref{fig:food-record-example} serves a different purpose. Its gallery
image and contrasting records were selected for visual clarity without inspecting predictions.
They illustrate how record selection changes a dietary decision, rather than document a scored
task instance or observed model error. The overview similarly depicts the resource and task
formats, not model predictions.

\subsection{Qualitative Dataset Examples}
\label{sec:app-dataset-examples}
\begin{figure}[t]
\centering
\begin{minipage}[t]{0.48\textwidth}
\centering\small
\textbf{A. A food-specific record contrast}\par\smallskip
\includegraphics[width=0.94\linewidth,height=32mm,keepaspectratio]{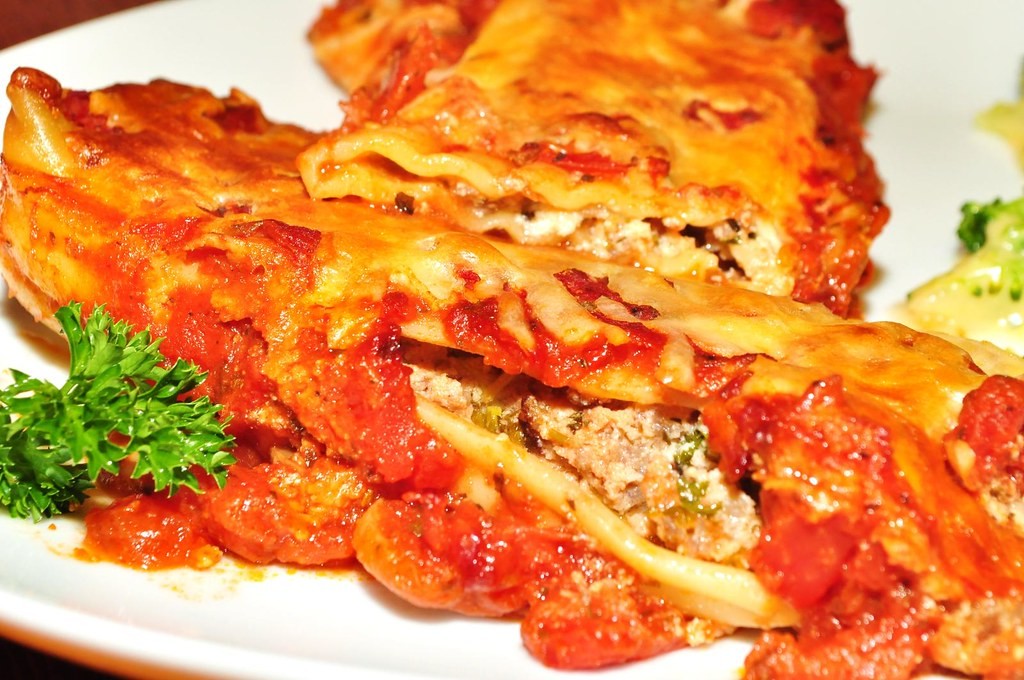}\par\smallskip
Manicotti: 393 mg sodium per 100 g.\par
Alternative meatless lasagna: 284 mg.
\end{minipage}\hfill
\begin{minipage}[t]{0.48\textwidth}
\centering\small
\textbf{B. One food within a mixed plate}\par\smallskip
\includegraphics[width=0.94\linewidth,height=32mm,keepaspectratio]{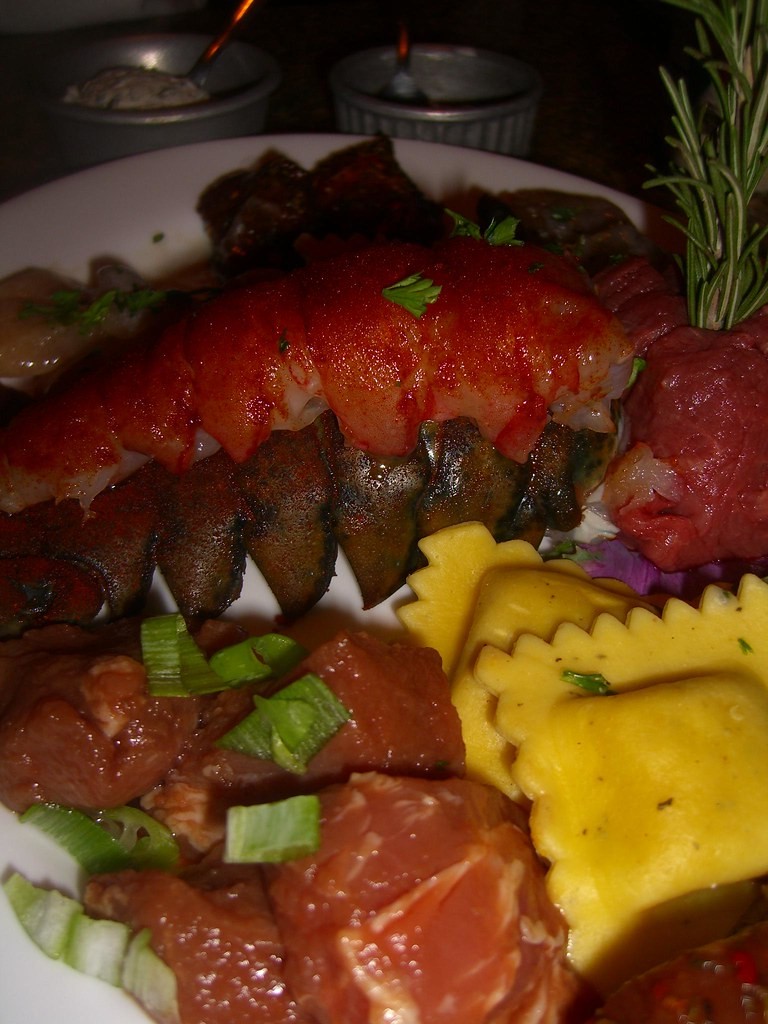}\par\smallskip
Cooked lobster: 88 kcal per 100 g.\par
Alternative fried scallops: 211 kcal.
\end{minipage}
\par\medskip
\begin{minipage}[t]{0.48\textwidth}
\centering\small
\textbf{C. A category is not a nutrient record}\par\smallskip
\includegraphics[width=0.94\linewidth,height=32mm,keepaspectratio]{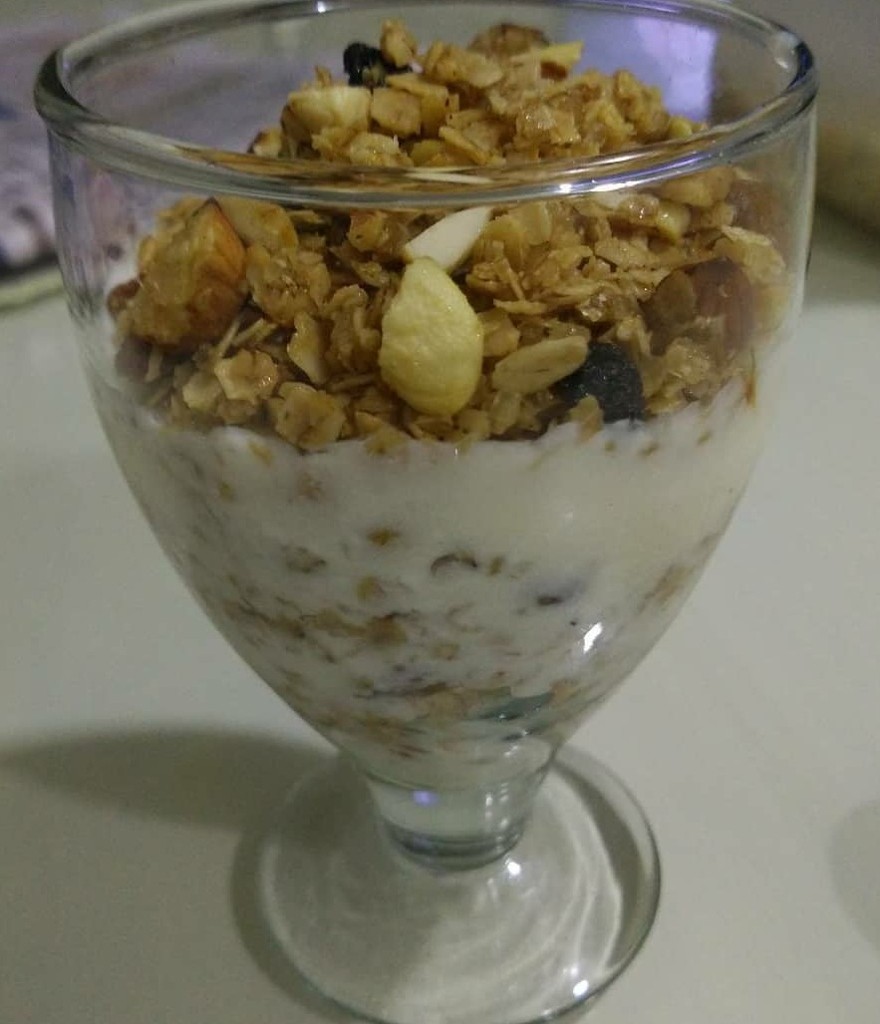}\par\smallskip
Homemade granola: 26 mg sodium per 100 g.\par
Alternative wheat-flake cereal: 718 mg.
\end{minipage}\hfill
\begin{minipage}[t]{0.48\textwidth}
\centering\small
\textbf{D. A visible image--record conflict}\par\smallskip
\includegraphics[width=0.94\linewidth,height=32mm,keepaspectratio]{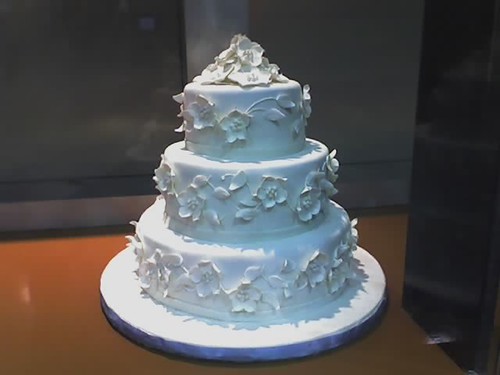}\par\smallskip
Assigned cream cake without icing or topping:\par
500 mg sodium per 100 g; visible icing excluded.
\end{minipage}
\caption{Curated VQA illustrations. Values belong to the assigned and alternative food records,
not measurements of the depicted meals. A and C contrast similar records; B and D expose the
scope and limitations of the assigned record. Image credits appear below.}
\label{fig:case-overview}
\label{fig:cases-vqa-records}
\end{figure}

\begin{figure}[t]
\centering
\begin{minipage}[t]{0.48\textwidth}
\centering\small
\textbf{E. Person A: compliant and relevant}\par\smallskip
\includegraphics[width=0.94\linewidth,height=35mm,keepaspectratio]{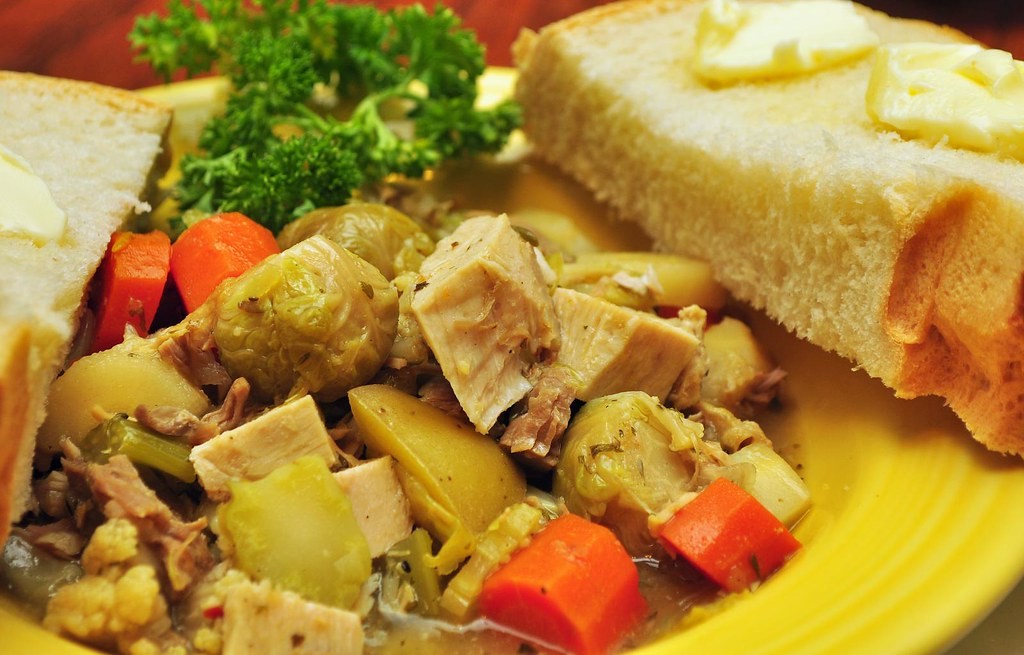}\par\smallskip
Assigned soup record passes the nutrient rule.\par
Its category is observed in later intake.
\end{minipage}\hfill
\begin{minipage}[t]{0.48\textwidth}
\centering\small
\textbf{F. Person B: compliant, relevance unobserved}\par\smallskip
\includegraphics[width=0.94\linewidth,height=35mm,keepaspectratio]{figures/dataset_cases/soup.jpg}\par\smallskip
Same assigned soup record and nutrient rule.\par
Its category is not observed in later intake.
\end{minipage}
\caption{Curated recommendation illustrations for two paired people with the same option images
and nutrient constraint. Compliance is unchanged while intake-derived relevance differs. These
panels illustrate option labels, not predicted rankings or a complete recommendation response.}
\label{fig:cases-recommendation-person}
\end{figure}

\begin{figure}[t]
\centering
\begin{minipage}[t]{0.48\textwidth}
\centering\small
\textbf{G. No valid replacement}\par\smallskip
\includegraphics[width=0.94\linewidth,height=35mm,keepaspectratio]{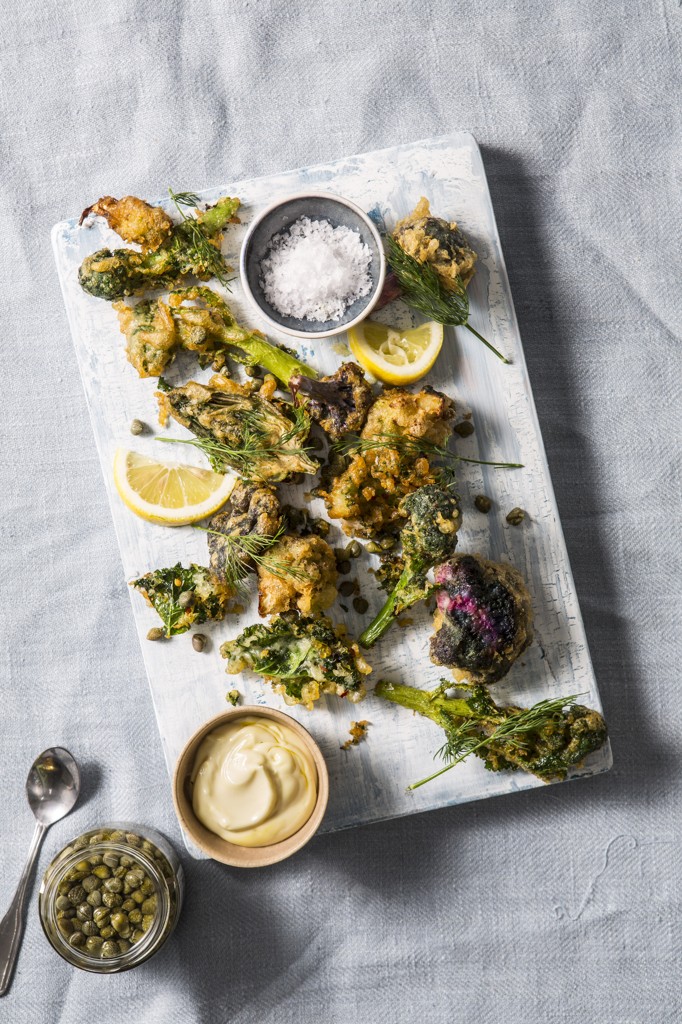}\par\smallskip
Assigned fried broccoli: 141 mg sodium per serving.\par
No candidate passes the complete rule: \texttt{none}.
\end{minipage}\hfill
\begin{minipage}[t]{0.48\textwidth}
\centering\small
\textbf{H. A valid swap with questionable grounding}\par\smallskip
\includegraphics[width=0.94\linewidth,height=35mm,keepaspectratio]{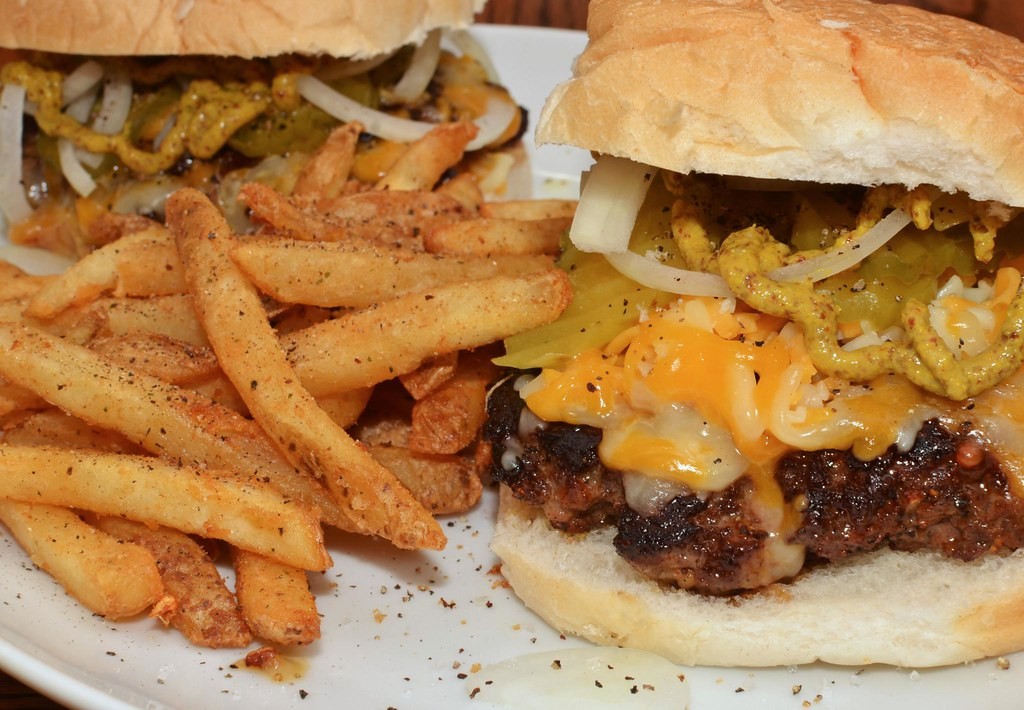}\par\smallskip
Assigned fried pickles: 372 mg sodium per serving.\par
Fried okra passes the complete replacement rule.
\end{minipage}
\caption{Curated substitution illustrations with candidate records and dietary constraints held
fixed. Changing the assigned source changes the correct response. Values are per database
serving; the gold decisions follow the complete rule, rather than a visual nutrient estimate.}
\label{fig:cases-substitution-records}
\end{figure}

Figures~\ref{fig:case-overview}--\ref{fig:cases-substitution-records} illustrate food-record
contrasts, person-dependent relevance, and substitution rules. We selected the cases for
contrasting decisions and visible image--record conflicts; they do not
estimate the prevalence of either pattern. The VQA and recommendation cases are validation-split illustrations, some selected after development error analysis. The
substitution cases come from a separate documentation set. None reports a model prediction or
score; observed test errors are summarized separately in Table~\ref{tab:app-error-types}.

The VQA cases separate record selection from annotation validity. In A and C, choosing the
similar alternative would change the nutrient answer even if its value were read correctly.
Recognizing a broad category such as breakfast cereal therefore does not resolve the record
contrast. In B, the assigned lobster record describes one food rather than the full plate or
the portion shown. Likewise, the granola record in C does not describe every ingredient of
the layered dessert. In D, the cake record explicitly excludes the visible icing. A correct
record-based answer thus need not establish the depicted food's nutrient content. These
conflicts qualify interpretation of the benchmark annotations without measuring their error rate.

The assigned chicken-or-turkey vegetable soup record satisfies the saturated-fat
constraint for both people, but its category appears only in person A's later intake. An
iced-cookie option also passes the constraint, while its category appears only in person B's
later intake. A rule-only choice of soup would therefore satisfy just one person's relevance
label. These observations provide incomplete implicit feedback: an unobserved category is
not evidence of dislike. We omit respondent identifiers, demographics, and intake histories.
Although the image source calls the dish turkey soup, its plated presentation does not
independently establish that assignment; the comparison concerns the assigned record.

In the substitution cases, the same candidate is valid for one source but not the other.
Fried okra contains 142.2 mg sodium per database serving;
it does not reduce the assigned fried broccoli's 141 mg. No supplied candidate satisfies
the complete rule for that source, so the answer is \texttt{none}. With the candidate pool
and constraints fixed, the assigned fried-pickle record instead contains 372 mg sodium,
and fried okra meets the target reduction, the bounds on other constrained nutrients, and
all personal limits. Its absolute sodium content alone cannot distinguish these decisions.
Panel H also exposes an image--record conflict: burgers and fries dominate the image, and
fried pickles are not clearly identifiable. The record-level answer remains valid under the
supplied annotation, while its visual grounding is questionable. Neither this conflict nor
the mixed preparation in G establishes the nutrient content of the entire depicted plate.

\paragraph{Image credits.}
We reuse the existing image files without cropping or aspect-ratio changes. Recorded credits
and licenses are: A, \href{https://www.flickr.com/photos/7927684@N03/4399223179}{jeffreyw},
\href{https://creativecommons.org/licenses/by/2.0/}{CC BY 2.0};
B, \href{https://www.flickr.com/photos/88774309@N00/2831758520}{permanently scatterbrained},
\href{https://creativecommons.org/licenses/by/2.0/}{CC BY 2.0};
C, \href{https://commons.wikimedia.org/wiki/File:GranolaDessert_HomeMade_Mumbai_(2).jpg}{Shah Naisargi},
\href{https://creativecommons.org/licenses/by-sa/4.0/}{CC BY-SA 4.0};
D, \href{https://www.flickr.com/photos/22481796@N00/1942864078}{carabou},
\href{https://creativecommons.org/licenses/by/2.0/}{CC BY 2.0};
E--F, \href{https://www.flickr.com/photos/7927684@N03/5219221176}{jeffreyw},
\href{https://creativecommons.org/licenses/by/2.0/}{CC BY 2.0};
G, \href{https://commons.wikimedia.org/wiki/File:Fried_Fritters_(Unsplash).jpg}{Stefan Johnson},
\href{https://creativecommons.org/publicdomain/zero/1.0/}{CC0};
H, \href{https://www.flickr.com/photos/7927684@N03/7564906260}{jeffreyw},
\href{https://creativecommons.org/licenses/by/2.0/}{CC BY 2.0}.

\FloatBarrier

\definecolor{promptcardblue}{HTML}{315FA8}
\definecolor{promptcardfill}{HTML}{EDF3FA}
\newsavebox{\promptcardbox}
\newenvironment{promptcard}[1]{%
\par\begingroup\setlength{\fboxsep}{6pt}\setlength{\fboxrule}{0.5pt}%
\begin{lrbox}{\promptcardbox}\begin{minipage}{\dimexpr\linewidth-2\fboxsep-2\fboxrule\relax}%
\colorbox{promptcardblue}{\parbox{\dimexpr\linewidth-2\fboxsep\relax}{\sffamily\bfseries\color{white}#1}}\par\smallskip
\footnotesize\raggedright\setlength{\parskip}{0pt}%
}{\end{minipage}\end{lrbox}\noindent\fcolorbox{promptcardblue}{promptcardfill}{\usebox{\promptcardbox}}\par\endgroup}
\newcommand{\promptfield}[1]{\par\vspace{3pt}\textbf{#1}\par\nobreak}
\newcommand{\promptanswer}[1]{\par\smallskip\begingroup\setlength{\fboxsep}{3pt}%
\noindent\fcolorbox{promptcardblue!45}{white}{\parbox{\dimexpr\linewidth-2\fboxsep-2\fboxrule\relax}{\footnotesize\ttfamily\raggedright#1}}\par\endgroup}

\clearpage
\section{Experiment Prompts}
\label{sec:app-prompts}
\subsection{Task 1: \tvqa}
\begin{figure}[H]
\centering
\begin{promptcard}{Visual Question Answering: direct-model prompt}
\promptfield{Input and candidate records}
You are given \textless{}image count\textgreater{} photograph(s), shown in order and labelled card \textless{}card labels\textgreater{}.\par
User context: none provided.\par
Knowledge records (JSON, shuffled; none is marked as belonging to any card): \textless{}shuffled candidate records, JSON\textgreater{}\par
\promptfield{Task instruction}
Task: \textless{}question\textgreater{}\par
\promptfield{Answer format: threshold question}
\promptanswer{Return JSON: \{"answer": "yes" or "no", "links": \{"\textless{}card\textgreater{}": \textless{}food code from the records or null\textgreater{}\}, "evidence": [record ids used]\}. If the supplied evidence is insufficient, return \{"status": "insufficient\_evidence", "links": \{...\}\}.\par}
\promptfield{Output requirements}
Put the "answer" field FIRST in the JSON object. Food codes are 8-digit integers copied from the records; write null when unsure.\par
Answer with the JSON object only, no explanation.\par
\promptfield{Question variant: threshold}
According to the supplied records, is the food shown on card A high in \textless{}nutrient\textgreater{} (at least \textless{}threshold and unit\textgreater{})?\par
\promptfield{Question variant: nutrient amount}
According to the supplied records, how much \textless{}nutrient\textgreater{} does the food on card A contain per 100 g, in \textless{}unit\textgreater{}?\par
\promptfield{Question variant: comparison}
According to the supplied records, which pictured food contains more \textless{}nutrient\textgreater{} per 100 g: card A or card B? Answer A, B, or tie if they are within \textless{}tolerance\textgreater{}\% of each other.\par
\promptfield{Alternative answer format: nutrient amount}
\promptanswer{Return JSON: \{"answer": \textless{}number\textgreater{}, "links": \{"\textless{}card\textgreater{}": \textless{}food code from the records or null\textgreater{}\}, "evidence": [record ids used]\}. If the evidence is insufficient, return \{"status": "insufficient\_evidence", "links": \{...\}\}.\par}
\promptfield{Alternative answer format: comparison}
\promptanswer{Return JSON: \{"answer": one of the listed options, "links": \{"\textless{}card\textgreater{}": \textless{}food code from the records or null\textgreater{}\}, "evidence": [record ids used]\}. If the evidence is insufficient, return \{"status": "insufficient\_evidence", "links": \{...\}\}.\par}
\end{promptcard}
\caption{VQA prompt, question variants, and required answer formats.}
\label{fig:prompt-vqa}
\end{figure}

The prompt templates specify the evidence each system receives and the answer the evaluator expects.
They make the distinction between grounding and subsequent record use inspectable: direct models
infer foods from images, whereas Graph-RAG readers receive predicted matches and retrieved evidence.
We reproduce the task-level prompts and reader instructions used by the evaluation code.

Angle brackets replace instance-dependent fields, including images, records, nutrient values, and
personal context; a single placeholder stands for repeated record lists. Quoted instructions retain
the implementation's wording. Line wrapping, bold field labels, and blue titles organize the
display without changing model input, while the model interface handles chat serialization.

Figure~\ref{fig:prompt-vqa} shows why VQA requires both grounding and record reading. Candidate
records contain the nutrient evidence but are not assigned to image cards, so the system must
establish that association before answering. The figure includes one fixed phrasing for each
question type. Amount questions specify the unit and basis; amount and comparison questions
replace the threshold answer instruction with their corresponding output formats.

\clearpage
\subsection{Task 2: \trec}
\begin{figure}[H]
\centering
\begin{promptcard}{Personalized Food Recommendation: direct-model prompt}
\promptfield{Input and personal context}
You are given 8 photographs, shown in order and labelled option A, B, C, D, E, F, G, H.\par
Person profile: \textless{}field\textgreater{} \textless{}value\textgreater{}.\par
Conditions: \textless{}condition\textgreater{}.\par
Food rules (an option that breaks any rule is never suitable):\par
  - \textless{}condition\textgreater{}: the option fails if its \textless{}nutrient\textgreater{} is at least \textless{}limit\textgreater{} \textless{}unit\textgreater{} per \textless{}serving basis\textgreater{}\par
Day-1 intake history (foods this person ate on the first recall day; graph entities): \textless{}food name\textgreater{} (food code \textless{}food code\textgreater{}).\par
\promptfield{Task instruction}
Task: These are 8 food options (photos A, B, C, D, E, F, G, H). Using this person's profile, conditions, the stated food rules and their day-1 intake history in the graph, rank all 8 options from most to least suitable for this person. An option that breaks one of the person's food rules is never suitable. Return every slot exactly once.\par
\promptfield{Answer format}
\promptanswer{Return JSON: \{"ranking": ["\textless{}slot\textgreater{}", ... all 8 slots, best first]\}\par}
\promptfield{Output requirements}
Put the "ranking" field FIRST in the JSON object and list the option letters only.\par
Answer with the JSON object only, no explanation.\par
\end{promptcard}
\caption{Recommendation prompt with dietary constraints and intake history.}
\label{fig:prompt-recommendation}
\end{figure}
\subsection{Task 3: \tsub}
\begin{figure}[H]
\centering
\begin{promptcard}{Condition-Aware Food Substitution: direct-model prompt}
\promptfield{Input, goal, and food records}
You are shown one food image.\par
Here is a food I eat. \textless{}goal sentence\textgreater{} My limits per serving: \textless{}nutrient\textgreater{} under \textless{}limit\textgreater{} \textless{}unit\textgreater{}.\par
The food in the image is one of these two: \textless{}source records\textgreater{}\par
Here are eight foods of the same kind with their facts: \textless{}candidate records\textgreater{}\par
\promptfield{Substitution rule and task instruction}
A better swap is a food of the same kind with at least 20\% less \textless{}goal nutrient\textgreater{} per serving than the food in the image, no other listed nutrient (energy, carbohydrate, saturated fat, sodium) more than 25\% higher per serving, and within all your limits. Which one is a better swap by this rule? Answer with one candidate id (c1-c8), or none if no candidate qualifies.\par
\promptfield{Answer format}
\promptanswer{Answer with a JSON object only: \{"answer": "\textless{}c1..c8 or none\textgreater{}"\}\par}
\promptfield{Source and candidate record format}
\textless{}entry\textgreater{}: \textless{}food name\textgreater{} (food code \textless{}food code\textgreater{}) | family: \textless{}family\textgreater{} | serving: \textless{}serving mass\textgreater{} g | per serving: energy \textless{}value\textgreater{} \textless{}unit\textgreater{}, carbohydrate \textless{}value\textgreater{} \textless{}unit\textgreater{}, total sugars \textless{}value\textgreater{} \textless{}unit\textgreater{}, saturated fat \textless{}value\textgreater{} \textless{}unit\textgreater{}, sodium \textless{}value\textgreater{} \textless{}unit\textgreater{} | ingredients: \textless{}ingredients\textgreater{}\par
\end{promptcard}
\caption{Substitution prompt and the shared food-record format.}
\label{fig:prompt-substitution}
\end{figure}

The recommendation prompt supplies personal context, explicit limits, and first-recall intake while
withholding option names and nutrient records. The system must therefore infer which foods the
images depict before checking compliance and ranking them for the person. Rule and history lines
repeat for each applicable limit and recorded food; intake history provides context for relevance
rather than changing the stated nutrient rules.

The substitution prompt supplies both possible source records and the replacement candidates.
Identifying the depicted source determines which nutrient values enter the replacement rule.
The goal and limits vary by instance, while the rule and answer instruction remain fixed.
Figure~\ref{fig:prompt-substitution} shows the shared record format: source entries use source
identifiers and replacements use candidate identifiers.

\clearpage
\subsection{Input Controls and Image Matching}
\begin{figure}[H]
\centering
\begin{promptcard}{Input controls and image matching}
\promptfield{Substitution: replacement goal when the goal is removed}
I want to cut one of these nutrients: energy, carbohydrate, saturated fat or sodium.\par
\promptfield{Substitution: opening for text-only input}
No image is shown to you.\par
\promptfield{VLM matcher: input and candidates}
You are given \textless{}image count\textgreater{} photograph(s) labelled card \textless{}card labels\textgreater{}. Candidate foods (JSON): \textless{}candidate names and food codes, JSON\textgreater{}\par
\promptfield{VLM matcher: task and answer format}
\promptanswer{For each card, which candidate food code does the photograph show? Return JSON only: \{"links": \{"\textless{}card\textgreater{}": \textless{}food code or null\textgreater{}\}\}.\par}
\end{promptcard}
\caption{Control-specific instructions and the VLM matching prompt.}
\label{fig:prompt-controls}
\end{figure}
\subsection{Graph-RAG Readers}
\begin{figure}[H]
\centering
\begin{promptcard}{Graph-RAG: common answer interface}
\promptfield{Image-to-record matching context}
Each card is a photograph the text model cannot see; it was linked to a food node by image search over the food gallery (may be wrong):\par
- card A -\textgreater{} \textless{}matched food name and code\textgreater{}\par
\promptfield{Retrieved evidence}
Knowledge graph facts (1-hop neighbourhood of the linked food nodes), as (head; relation; tail) triples: \textless{}retrieved triples\textgreater{}\par
\promptfield{Question}
Question: \textless{}question\textgreater{}\par
\promptfield{Answer format}
\promptanswer{Return only a JSON object of the form \{"answer": "yes" or "no"\}. If the evidence cannot decide the question, return \{"status": "insufficient\_evidence"\}.\par}
\promptfield{KAPING: alternative evidence header}
Below are facts in the form of triples meaningful to answer the question.\par
\end{promptcard}
\caption{Graph-RAG reader prompt, shown for binary VQA.}
\label{fig:prompt-rag}
\end{figure}

The controls change available evidence while retaining the task instruction, allowing comparisons
with the standard input (Table~\ref{tab:conditions}). Figure~\ref{fig:prompt-controls} gives the
substitution-specific changes and the VLM matching instruction. Image removal substitutes gray
canvases and explicitly states their absence; true-food controls supply correct identities and
additionally supply option records for recommendation. Image-order controls retain the original
card labels. These comparisons diagnose sensitivity to input conditions rather than isolate
disjoint error causes.

The matcher is asked to identify foods without the dietary decision instruction. VQA and
substitution restrict its candidates to the supplied possible foods; recommendation matches each
option separately against retrieved gallery candidates. This interface keeps image-to-record
selection distinct from the reader's use of nutrient and respondent evidence.

Figure~\ref{fig:prompt-rag} shows the common Graph-RAG interface for VQA. The reader receives
predicted identities and retrieved triples, so an answer can reflect both matching errors and
errors in using the evidence. Recommendation additionally retrieves respondent evidence and
requests an option ranking; substitution requests a candidate identifier or \texttt{none}.
KAPING ranks available triples by question relevance and uses the alternative evidence header
shown in the figure.

\clearpage
\begin{figure}[H]
\centering
\begin{promptcard}{Zero-shot CoT: reasoning and answer extraction}
\promptfield{Reasoning instruction: appended to the first user turn}
Reason step by step, then give the final answer.\par
\promptfield{Assistant prefill}
Let's think step by step.\par
\promptfield{Answer-extraction continuation}
\promptanswer{Therefore, the final answer, as a JSON object of the form \textless{}task-specific JSON schema\textgreater{}, is: \textless{}answer prefill\textgreater{}\par}
\promptfield{Retry instruction when answer extraction fails}
Give only the final answer now. \textless{}task-specific JSON instruction\textgreater{}\par
\end{promptcard}
\caption{CoT instructions, continuation, and answer-extraction retry.}
\label{fig:prompt-cot}
\end{figure}
\begin{figure}[H]
\centering
\begin{promptcard}{Think-on-Graph: evidence selection and sufficiency}
\promptfield{Stage 1: relation selection}
Please retrieve up to 3 relations that contribute to the question and rate their contribution on a scale from 0 to 1 (the scores sum to 1).\par
\textless{}matching context\textgreater{}\par
Question: \textless{}question\textgreater{}\par
Topic entity: \textless{}entity\textgreater{}\par
Available relations: \textless{}relation list\textgreater{}\par
\promptfield{Required relation-selection output}
\promptanswer{Return only a JSON object: \{"relations": [\{"relation": "\textless{}copied exactly\textgreater{}", "score": \textless{}0-1\textgreater{}\}]\}\par}
\promptfield{Stage 2: entity selection}
Score the candidate entities by their contribution to the question (scores sum to 1) and keep up to 3.\par
\textless{}matching context\textgreater{}\par
Question: \textless{}question\textgreater{}\par
Relation: (\textless{}entity\textgreater{}; \textless{}relation\textgreater{}; ?)\par
Candidate entities: \textless{}entity list\textgreater{}\par
\promptfield{Required entity-selection output}
\promptanswer{Return only a JSON object: \{"entities": [\{"entity": "\textless{}copied exactly\textgreater{}", "score": \textless{}0-1\textgreater{}\}]\}\par}
\promptfield{Stage 3: evidence sufficiency}
\textless{}matching context\textgreater{}\par
Knowledge triples retrieved so far: \textless{}retrieved triples\textgreater{}\par
Question: \textless{}question\textgreater{}\par
\promptfield{Required sufficiency output}
\promptanswer{Are these triples sufficient to answer the question? Answer only \{"sufficient": "yes"\} or \{"sufficient": "no"\}.\par}
\end{promptcard}
\caption{Think-on-Graph prompts for relation selection, entity selection, and sufficiency.}
\label{fig:prompt-tog}
\end{figure}

Zero-shot CoT adds the instruction and assistant prefill in Figure~\ref{fig:prompt-cot} to the
matching context, records, question, and personal context when applicable. Answer extraction
retains the generated reasoning rather than answering from a new evidence set. If the answer
cannot be parsed, an additional user turn requests an answer over that same reasoning.

Think-on-Graph (Figure~\ref{fig:prompt-tog}) selects relations and entities before assessing
whether the accumulated evidence suffices. The matching context and question remain available
through these stages. For recommendation, the question also contains the full public profile,
nutrient rules, and first-recall history. The final reader receives this context and the selected
triples with the task-specific answer instruction. Assistant turns use the corresponding JSON opening as a prefill.
Appendix~\ref{sec:app-protocol} specifies generation settings and output handling.

\begin{table}[t]
\centering\tabfont\setlength{\tabcolsep}{3pt}
\begin{adjustbox}{max width=\linewidth}
\begin{tabular}{@{}lcccc@{}}
\toprule
& & \multicolumn{3}{c}{\textbf{Share of all answers wrong with}} \\
\cmidrule(lr){3-5}
\textbf{System} & \textbf{Right matches} & \textbf{a wrong match} & \textbf{right matches} & \textbf{no stated match} \\
\midrule
\rowcolor{groupgray}\multicolumn{5}{@{}l}{\textit{Match each image to a food, then apply the rules}} \\
DINOv2 match + rules & 0.722\pmse{0.024} & 0.278\pmse{0.024} & 0.000 & 0.000 \\
SigLIP2 zero-shot match + rules & 0.922\pmse{0.014} & 0.078\pmse{0.014} & 0.000 & 0.000 \\
SigLIP2 gallery match + rules & 0.684\pmse{0.023} & 0.316\pmse{0.023} & 0.000 & 0.000 \\
SigLIP2 fused match + rules & 0.827\pmse{0.019} & 0.173\pmse{0.019} & 0.000 & 0.000 \\
Qwen3-VL-8B match + rules & 0.838\pmse{0.015} & 0.086\pmse{0.011} & 0.000 & 0.076\pmse{0.011} \\
\rowcolor{groupgray}\multicolumn{5}{@{}l}{\textit{Graph-RAG with SigLIP2 matches (point estimates)}} \\
Retrieved records + Llama-3.1-8B & 0.922 & 0.073 & 0.105 & 0.000 \\
Retrieved records + Qwen3-8B & 0.922 & 0.076 & 0.128 & 0.000 \\
Retrieved records + Qwen3-8B (R) & 0.922 & 0.078 & 0.001 & 0.000 \\
KAPING + Llama-3.1-8B & 0.922 & 0.076 & 0.085 & 0.000 \\
KAPING + Qwen3-8B & 0.922 & 0.076 & 0.086 & 0.000 \\
KAPING + Qwen3-8B (R) & 0.922 & 0.078 & 0.001 & 0.000 \\
Zero-shot CoT + Llama-3.1-8B & 0.922 & 0.066 & 0.098 & 0.000 \\
Zero-shot CoT + Qwen3-8B & 0.922 & 0.077 & 0.006 & 0.000 \\
Think-on-Graph + Llama-3.1-8B & 0.922 & 0.076 & 0.056 & 0.000 \\
Think-on-Graph + Qwen3-8B & 0.922 & 0.077 & 0.040 & 0.000 \\
\rowcolor{groupgray}\multicolumn{5}{@{}l}{\textit{Vision--language models answering directly}} \\
Qwen3-VL-2B & 0.455\pmse{0.013} & 0.211\pmse{0.010} & 0.045\pmse{0.004} & 0.159\pmse{0.008} \\
Qwen3-VL-4B & 0.644\pmse{0.013} & 0.167\pmse{0.010} & 0.067\pmse{0.005} & 0.085\pmse{0.007} \\
Qwen3-VL-8B & 0.712\pmse{0.013} & 0.123\pmse{0.009} & 0.079\pmse{0.005} & 0.071\pmse{0.005} \\
InternVL3.5-8B & 0.653\pmse{0.013} & 0.188\pmse{0.010} & 0.123\pmse{0.007} & 0.068\pmse{0.006} \\
GLM-4.1V-9B (R) & 0.677\pmse{0.013} & 0.085\pmse{0.008} & 0.001\pmse{0.001} & 0.048\pmse{0.006} \\
MiMo-VL-7B-RL & 0.741\pmse{0.014} & 0.080\pmse{0.008} & 0.006\pmse{0.001} & 0.078\pmse{0.009} \\
Qwen3-VL-8B (R) & 0.872\pmse{0.012} & 0.068\pmse{0.010} & 0.001\pmse{0.001} & 0.058\pmse{0.008} \\
\bottomrule
\end{tabular}

\end{adjustbox}
\caption{VQA test errors by recorded grounding outcome, pooling the listed order and the seeded
reordering. Error columns use all answers as the denominator. $\pm$: bootstrap SE. Graph-RAG rows use the
SigLIP2 matches of Table~\ref{tab:t1-main} and give point estimates (Appendix~\ref{sec:app-why-t1}).}
\label{tab:app-error-types}
\end{table}

\end{document}